\documentclass[sigconf]{acmart}

\usepackage{amsmath,amsfonts}
\usepackage{algorithm}
\usepackage{algorithmic}
\usepackage{graphicx}
\usepackage{textcomp}
\usepackage{xcolor}

\usepackage{graphicx}
\usepackage{amsmath}
\usepackage{booktabs}

\usepackage{url}            
\usepackage{booktabs}       
\usepackage{amsfonts}       
\usepackage{nicefrac}       
\usepackage{microtype}      
\usepackage{xcolor}         

\usepackage{nccmath}  
\usepackage{bbm}
\usepackage{amsmath}
\usepackage{mathtools}
\usepackage{amsthm}

\usepackage{amsfonts}

\usepackage[flushleft]{threeparttable} 
\usepackage{multirow}

\usepackage{enumitem}
\usepackage{wrapfig}

\usepackage{tikz}
\newcommand{\ballnumber}[1]{\tikz[baseline=(myanchor.base)] \node[circle,fill=.,inner sep=1pt] (myanchor) {\color{-.}\bfseries\footnotesize #1};}

\newcommand{\tf}[1]{\mathbf{#1}}
\newcommand{\R}{\mathbb{R}}

\definecolor{Note_color}{rgb}{0.0, 0.0, 0.0}

\newcommand{\YL}[1]{{\color{black}{\textbf{}#1}}}
\newcommand{\cw}[1]{{\color{black}{#1}}}

\newcommand*\dif{\mathop{}\!\mathrm{d}}

\usepackage{layout}
\usepackage{pifont}
\newcommand\mynuma[1]{\ifcase#1 \or \ding{172}\or \ding{173}\or
  \ding{174}\or \ding{175}\or \ding{176}\or \ding{177}%
  \or \ding{178}\or \ding{179}\or \ding{180}\or \ding{181}\else *\fi\relax}

\newcommand\mynumb[1]{\ifcase#1 \or \ding{182}\or \ding{183}\or
  \ding{184}\or \ding{185}\or \ding{186}\or \ding{187}%
  \or \ding{188}\or \ding{189}\or \ding{190}\or \ding{191}\else *\fi\relax}

\newcommand{\METHOD}{Gen-NeRF} 

\AtBeginDocument{%
  \providecommand\BibTeX{{%
    \normalfont B\kern-0.5em{\scshape i\kern-0.25em b}\kern-0.8em\TeX}}}

\setcopyright{acmcopyright}
\copyrightyear{2023}
\acmYear{2023}
\acmDOI{3579371.3589109}

\copyrightyear{2023}
\acmYear{2023}
\setcopyright{rightsretained}
\acmConference[ISCA '23]{Proceedings of the 50th Annual International
Symposium on Computer Architecture}{June 17--21, 2023}{Orlando, FL, USA}
\acmBooktitle{Proceedings of the 50th Annual International Symposium on
Computer Architecture (ISCA '23), June 17--21, 2023, Orlando, FL, USA}
\acmDOI{10.1145/3579371.3589109} 
\acmISBN{979-8-4007-0095-8/23/06}

\begin{document}

\title{Gen-NeRF: Efficient and Generalizable Neural Radiance Fields via Algorithm-Hardware Co-Design}

\author{Yonggan Fu}
\authornote{Equal contribution.}
\affiliation{%
  \institution{Georgia Institute of Technology}
  \streetaddress{North Avenue Atlanta}
  \city{Atlanta}
  \state{Georgia}
  \country{USA}
  \postcode{30332}
}
\email{yfu314@gatech.edu}

\author{Zhifan Ye}
\authornotemark[1]
\affiliation{%
  \institution{Georgia Institute of Technology}
  \streetaddress{North Avenue Atlanta}
  \city{Atlanta}
  \state{Georgia}
  \country{USA}
  \postcode{30332}
}
\email{zye327@gatech.edu}

\author{Jiayi Yuan}
\affiliation{%
  \institution{Rice University}
  \streetaddress{6100 Main St}
  \city{Houston}
  \state{Texas}
  \country{USA}
  \postcode{77005}}
\email{jy101@rice.edu}

\author{Shunyao Zhang}
\affiliation{%
  \institution{Rice University}
  \streetaddress{6100 Main St}
  \city{Houston}
  \state{Texas}
  \country{USA}
  \postcode{77005}}
\email{sz74@rice.edu}

\author{Sixu Li}
\affiliation{%
  \institution{Georgia Institute of Technology}
  \streetaddress{North Avenue Atlanta}
  \city{Atlanta}
  \state{Georgia}
  \country{USA}
  \postcode{30332}
}
\email{sli941@gatech.edu}

\author{Haoran You}
\affiliation{%
  \institution{Georgia Institute of Technology}
  \streetaddress{North Avenue Atlanta}
  \city{Atlanta}
  \state{Georgia}
  \country{USA}
  \postcode{30332}
}
\email{hyou37@gatech.edu}

\author{Yingyan (Celine) Lin}
\affiliation{%
  \institution{Georgia Institute of Technology}
  \streetaddress{North Avenue Atlanta}
  \city{Atlanta}
  \state{Georgia}
  \country{USA}
  \postcode{30332}
}
\email{celine.lin@gatech.edu}


\begin{abstract}

Novel view synthesis is an essential functionality for enabling immersive experiences in various Augmented- and Virtual-Reality (AR/VR) applications, for which Neural Radiance Field (NeRF) has emerged as the state-of-the-art (SOTA) technique. In particular, generalizable NeRFs have gained increasing popularity thanks to their cross-scene generalization capability, which enables NeRFs to be instantly serviceable for new scenes without per-scene training. Despite their promise, generalizable NeRFs aggravate the prohibitive complexity of NeRFs due to their required extra memory accesses needed to acquire scene features, causing NeRFs’ ray marching process to be memory-bounded. To tackle this dilemma, existing sparsity-exploitation techniques for NeRFs fall short, because they require knowledge of the sparsity distribution of the target 3D scene which is unknown when generalizing NeRFs to a new scene.

To this end, we propose Gen-NeRF, an algorithm-hardware co-design framework dedicated to generalizable NeRF acceleration, which aims to win both rendering efficiency and generalization capability in NeRFs. To the best of our knowledge, Gen-NeRF is the first to enable real-time generalizable NeRFs, demonstrating a promising NeRF solution for next-generation AR/VR devices. On the algorithm side, Gen-NeRF integrates a coarse-then-focus sampling strategy, leveraging the fact that different regions of a 3D scene contribute differently to the rendered pixels depending on where the objects are located in the scene, to enable sparse yet effective sampling. In addition, Gen-NeRF replaces the ray transformer, which is generally included in SOTA generalizable NeRFs to enhance density estimation, with a novel Ray-Mixer module to reduce workload heterogeneity. On the hardware side, Gen-NeRF highlights an accelerator micro-architecture dedicated to accelerating the resulting model workloads from our Gen-NeRF algorithm to maximize the data reuse opportunities among different rays by making use of their epipolar geometric relationship. Furthermore, our Gen-NeRF accelerator features a customized dataflow to enhance data locality during point-to-hardware mapping and an optimized scene feature storage strategy to minimize memory bank conflicts across camera rays of NeRFs. Extensive experiments validate the effectiveness of our proposed Gen-NeRF framework in enabling real-time and generalizable novel view synthesis.

\end{abstract}





\maketitle

\section{Introduction}
\label{sec:intro}

The booming Augmented- and Virtual-Reality (AR/VR) market has motivated a tremendously increased demand for immersive AR/VR experiences~\cite{mystakidis2022Metaverse}. To enable a truly immersive interactive AR/VR experience, novel view synthesis which can generate a novel view of a scene given only sparsely sampled views, is one key enabler~\cite{bian2016framework,zhao2020deja,fassi2016vr,farshid2018go}. 
Hence, substantial advances have been made in novel view synthesis to enhance the 3D scene representation~\cite{park2019deepsdf,mildenhall2020nerf,riegler2020free,jiang2020local,genova2020local}. Among them, 
Neural Radiance Field (NeRF)~\cite{mildenhall2020nerf}, which encodes a continuous volume representation of the density and view-dependent color of a target scene, has gained increasing popularity thanks to its state-of-the-art (SOTA) rendering quality for photorealistic novel views.

Despite NeRFs' impressive rendering quality, enabling real-time rendering of NeRFs on resource-constrained AR/VR devices, which is highly desirable for numerous AR/VR applications, is still particularly challenging due to the following bottlenecks: (1) Vanilla NeRFs require per-scene optimization and thus cannot be effectively generalized to a new scene specified by a user; (2) The volume rendering process in NeRFs has a prohibitive complexity of $H \times W \times P$, where $H$ and $W$ denote the height and width of the rendered image, respectively, and $P$ is the number of sampled points along each camera ray. 
Such a cubic complexity results in a throughput of $\leq$0.1 frame-per-second (FPS) even on an NVIDIA desktop GPU~\cite{hedman2021baking}.

To tackle the aforementioned bottleneck-(1), generalizable NeRFs \cite{yu2021pixelnerf,wang2021ibrnet,chen2021mvsnerf,liu2022neural} have become the mainstream solution for improving NeRFs' generalization capability. Specifically, their key spirit is to condition vanilla NeRFs on a set of source views by extracting scene features from those source views, which are then fed into vanilla NeRFs as inputs. By doing so, generalizable NeRFs 
eliminate the need for per-scene optimization and thus hold the promise of being the most feasible NeRF solution for commercial AR/VR applications.

However, the boosted generalization capability of generalizable NeRFs comes at the cost of aggravating the aforementioned bottleneck-(2), posing new challenges for achieving real-time NeRFs. 
First, conditioning NeRFs on the source views mentioned above requires extra memory accesses to fetch scene features, which can make NeRFs' ray marching process memory-bounded and thus cause hardware under-utilization issues. In particular, to infer the density and color of each sampled point in a 3D scene, generalizable NeRFs project the sampled point onto the image planes of $S$ different source views and then apply the resulting $D$-dimension scene features of the projection point to the inputs of vanilla NeRFs. 
As such, the total number of memory accesses for acquiring the scene features becomes $H \times W \times P \times S \times D$, which can result in significant latency overhead as profiled in Sec.~\ref{sec:profile}.

The second challenge is that while it is natural to consider using SOTA sparsity-exploitation techniques for NeRFs to boost the acceleration efficiency of generalizable NeRFs, these techniques rely on the knowledge of the spatial sparsity distribution in the target 3D scene, which is unknown for a new scene. 
Therefore, these techniques are not applicable to generalizable NeRFs because the spatial distributions of different scenes may vary significantly, thus making it infeasible to predict and utilize the spatial sparsity of new scenes. Third, extra ray transformers~\cite{wang2021ibrnet,reizenstein2021common,wang2022attention} are often introduced in SOTA model structures of generalizable NeRFs for more accurately predicting the densities of new scenes that have a complex geometry~\cite{wang2021ibrnet}. These additional transformer modules with attention operations~\cite{vaswani2017attention} increase the workload heterogeneity of generalizable NeRFs, further challenging their execution efficiency.

To tackle the aforementioned challenges of enabling real-time generalizable NeRFs, we first identify opportunities unique to generalizable NeRFs that can boost their achievable acceleration efficiency and then 
develop an algorithm-hardware co-design framework, dubbed \METHOD{}, which to the best of our knowledge is the first to achieve real-time efficiency of generalizable NeRFs. 

On the algorithm side, the key insight that motivates our work is that the contributions of different regions in a 3D scene to the rendered pixels can vary depending on the location of objects, offering an opportunity for sparse sampling. Specifically, sampled points in empty or occluded regions of the scene contribute less to the rendered pixels and thus fewer sampled points are needed in such regions of the target 3D scene. To leverage this, our \METHOD{} algorithm integrates a coarse-then-focus sampling scheme to enable sparse yet effective sampling. 
Additionally, our \METHOD{} algorithm further reduces the workload heterogeneity in generalizable NeRF models by introducing an MLP-based module, dubbed Ray-Mixer, to replace the ray transformer~\cite{wang2021ibrnet,reizenstein2021common,wang2022attention} in SOTA generalizable NeRFs. The advantage of doing this is that the Ray-Mixer module can maintain the capability of accurately estimating the density of the former while enabling the reuse of the computing units that are dedicated to generalizable NeRFs' MLPs needed for implicitly encoding the continuous volume representation.

On the hardware side, we discover that there exist inherent opportunities for making use of the geometric relationships among different camera rays to reduce the required number of memory accesses for acquiring scene features of different source views. 
As such, we develop a dedicated accelerator to accelerate the resulting workloads from our \METHOD{} algorithm. In particular, our \METHOD{} accelerator leverages the epipolar geometric analysis~\cite{zhang1998determining} and highlights three components: (1) a customized dataflow that enhances data locality during point-to-hardware mapping. More specifically, we partition the points in the 3D scene into point patches that can be projected to the same or neighboring regions on the image planes of different source views based on their epipolar geometric relationships, thus enhancing scene feature reuses; 
(2) an optimized scene feature storage strategy for avoiding memory bank conflicts when loading scene features of different rays;
(3) a customized ray marching micro-architecture that accelerates \METHOD{}'s algorithm by orchestrating the coarse and focused sampling processes and features a run-time workload scheduler to efficiently execute the above 3D-point-patch partition at run-time.
Finally, we summarize our contributions as follows:

\begin{figure*}[!t]
\centering
\includegraphics[width=0.9\linewidth]{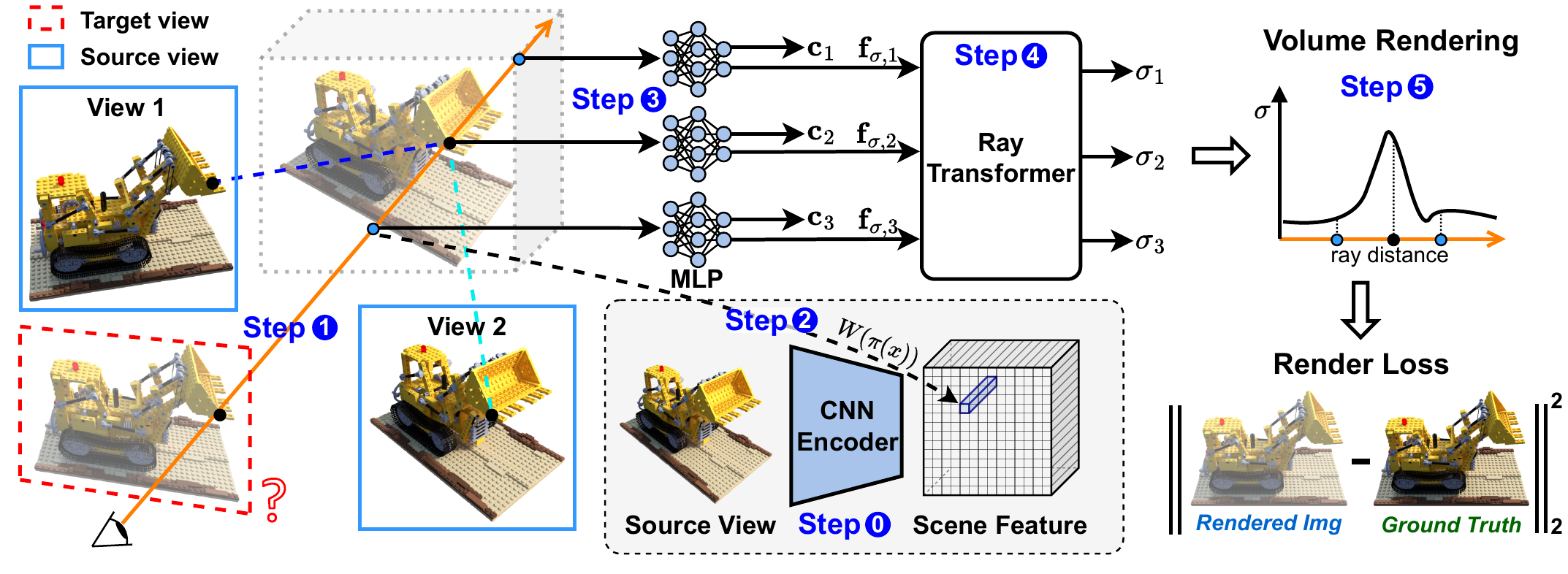}
\vspace{-1.2em}
\caption{
Visualizing the typical execution pipeline of generalizable NeRFs~\cite{wang2021ibrnet,reizenstein2021common,wang2022attention}, which condition NeRF on source views and enhance density estimation via a ray transformer. This illustration is modified from the visualization style of~\cite{wang2021ibrnet}.
}
\label{fig:workload}
\vspace{-1.3em}
\end{figure*}

\begin{itemize}[topsep=0.2em]
    \item We propose an algorithm-hardware co-design framework, dubbed \METHOD{}, which is the first to enable real-time generalizable NeRFs, offering a promising NeRF solution for next-generation AR/VR applications. Furthermore, the opportunities we identify can also shed light on future innovations for accelerating more diverse NeRF pipelines.

    \item On the algorithm side, \METHOD{} integrates a coarse-then-focus sampling strategy that leverages the fact that different regions in a 3D scene can feature diverse sparsity ratios depending on where the objects are located in the scene to enable sparse yet effective sampling. 
    In addition, \METHOD{} develops a novel Ray-Mixer module, which replaces the ray transformer that is generally included in SOTA generalizable NeRFs in order to enhance NeRFs' density estimation, aiming at reducing workload heterogeneity.

    \item On the hardware side, \METHOD{} highlights an accelerator micro-architecture dedicated to accelerating the resulting model workloads from our \METHOD{} algorithm to maximize the data reuse opportunities among different rays by making use of their epipolar geometric relationships. Furthermore, our \METHOD{} accelerator features a customized dataflow to enhance data locality during point-to-hardware mapping and an optimized scene feature storage strategy to minimize memory bank conflicts across camera rays.
    
    \item Extensive experiments validate the effectiveness of our \METHOD{} in enabling real-time and generalizable novel view synthesis, e.g., \METHOD{} achieves a 255.8$\times$ speed-up over the NVIDIA RTX 2080Ti GPU while maintaining a photorealistic rendering quality.
    
\end{itemize}

\section{Preliminaries and Analysis of Generalizable NeRF Workloads}
\label{sec:background}

\subsection{Preliminaries of NeRFs}
\label{sec:preliminary}

We first introduce vanilla NeRFs' rendering pipeline. 
To render a pixel corresponding to a camera ray that is emitted from the camera center and passes through this pixel, NeRF performs a ray marching process, i.e., it samples 3D points along the ray, estimates the color and density of each sampled point, and then composites the colors and densities of the sampled points to derive the pixel value.

Specifically, a camera ray can be parameterized as $\tf r(t) = \tf o + t \tf d$, with $\tf o \in \R^3$ denoting the ray origin (i.e., the camera center) and $\tf d \in \R^3$ denoting the ray unit direction vector, where $t \in [t_n, t_f]$ is the depth along the ray between the predefined near bound $t_n$ and far bound $t_f$.
To acquire the color and density of each sampled point given both its location in the 3D space $\tf x \in \R^3$ and view direction unit vector $\tf d \in \R^3$, a volumetric radiance field $f$ returns a differential density $\sigma$ and RGB color $\tf c$, i.e., ~\((\sigma, \tf c)\ = f(\tf x, \tf d)\). Next, the volume along the ray $\tf r$ can be rendered into a 2D pixel $\hat{\mathbf{C}}(\mathbf{r})$ via an integral over the colors of sampled points: 

\vspace{-0.1em}
\begin{equation}
     \hat{\mathbf{C}}(\mathbf{r}) = \int_{t_n}^{t_f} T(t) \sigma(t) \mathbf{c}(t) \dif t  
 \label{eq:rendering}
\end{equation}
\vspace{-1em}

\noindent where~$T(t) = \exp\left(- \int_{t_n}^{t} \sigma(s) \,\dif s \right)$ denotes the accumulated transmittance along the ray from $t_n$ to $t$, which refers to the probability that the ray travels from $t_n$ to $t$ without hitting any other particle and is to measure the occlusion effect.
In practice, the integral in Eq.~\ref{eq:rendering} is approximated with numerical quadrature by sampling $N$ points along each camera ray:

\begin{align}
\hat{\mathbf{C}}(\mathbf{r}) = \sum_{k=1}^N T_k\left(1-\exp(-\sigma_k \left(t_{k+1}-t_{k})\right)\right)\mathbf{c}_k
\label{eq:nerf_render}
\end{align}

\noindent where $T_k = \exp\left(-\sum_{j=1}^{k-1} \sigma_{j} (t_{j+1}-t_{j})\right)$. To train the volumetric radiance field $f$, a Mean-Square-Error (MSE) loss is applied between the rendered pixels and the ground truth pixels from all camera rays of the target view:

\begin{equation}
    \mathcal{L} = \sum_{\tf r \in \mathcal{R}} \left\lVert
   \hat{\tf C}(\tf r) - \tf C(\tf r) \right\rVert_2^2
    \label{eq:mseloss}
\end{equation}

\noindent where $\mathcal{R}$ is the set of all camera rays. 
As the inputs to vanilla NeRFs only include the scene-invariant point location $\tf x$ and view direction $\tf d$, it is difficult to generalize them across different scenes.

\subsection{The Pipeline of Generalizable NeRFs}
\label{sec:workload}

Generalizable NeRF variants~\cite{wang2021ibrnet,chen2021mvsnerf,reizenstein2021common,liu2022neural} enable cross-scene generalization via two modifications on top of vanilla NeRFs: (1) conditioning NeRFs on the source views of new scenes, i.e., given a limited number of observed source views of a new scene, the features extracted from those source views via a CNN encoder are used as scene priors and fed into a vanilla NeRF model as inputs, and (2) adopting a ray transformer on top of all the points across the same ray to enhance the density prediction.

As an example, we illustrate the execution pipeline of a representative generalizable NeRF called IBRNet~\cite{wang2021ibrnet} in Fig.~\ref{fig:workload}, which is the first to propose the two aforementioned modules and has served as a cornerstone for follow-up generalizable NeRF variants. Specifically, rendering a pixel in IBRNet involves the following steps: 
Step \ballnumber{0} calculates the \cw{2D feature maps} $\{\tf W_i\}_{i=1}^S $ from a total of $S$ source views $\{\tf I_i\}_{i=1}^S $ via a CNN encoder $E$, where $\tf W_i = E(\tf I_i)$ \cw{is a 3D tensor}. Note that this requires only a one-time effort for each new scene; Step \ballnumber{1} emits a ray $\tf r(t) = \tf o + t \tf d$ from the origin $\tf o$ along the view direction $\tf d$ to pass through the pixel to be rendered and sample \cw{3D points} $\{\tf x_k\}$ along the ray based on \cw{an ordered depth sequence $\{t_k\}$ sampled from} a certain distribution; 
Step \ballnumber{2} projects each sampled 3D point $\tf x_k$ to the image planes of source views \cw{with a project transformation $\pi$} and acquires the corresponding scene features, i.e., \cw{$\{\tf [W_i]_{\pi(\tf x_k)}\}_{i=1}^S$}; Step \ballnumber{3} applies the obtained scene features above to an MLP model $f$ to derive the color \YL{$\mathbf{c}_k$ and density feature $f_k^{\sigma}$} of each point; Step \ballnumber{4} feeds the density features of all sampled points along the ray into a ray transformer $T$ to acquire the predicted density $\sigma_k$ for each point; Step \ballnumber{5} performs volume rendering following Eq.~\ref{eq:nerf_render} to finally derive the rendered pixel. During training, the networks $E$, $f$, and $T$ are updated using the MSE loss in Eq.~\ref{eq:mseloss}.

\subsection{Profiling Results and Analysis}
\label{sec:profile}

\textbf{Setup.}
To understand the real-device efficiency of generalizable NeRFs,
we profile our adopted generalizable NeRF model, which is built on top of~\cite{wang2021ibrnet} as elaborated in Sec.~\ref{sec:evaluate_alg}, in terms of the latency breakdown for rendering one image with 10 source views and 196 points per ray, following~\cite{wang2021ibrnet}, on two devices, including a desktop GPU NVIDIA RTX 2080Ti and an edge GPU NVIDIA Jetson TX2, with a batch size of 4096 and 128 rays, respectively. 

\begin{figure}[!t]
\centering
\includegraphics[width=0.95\linewidth]{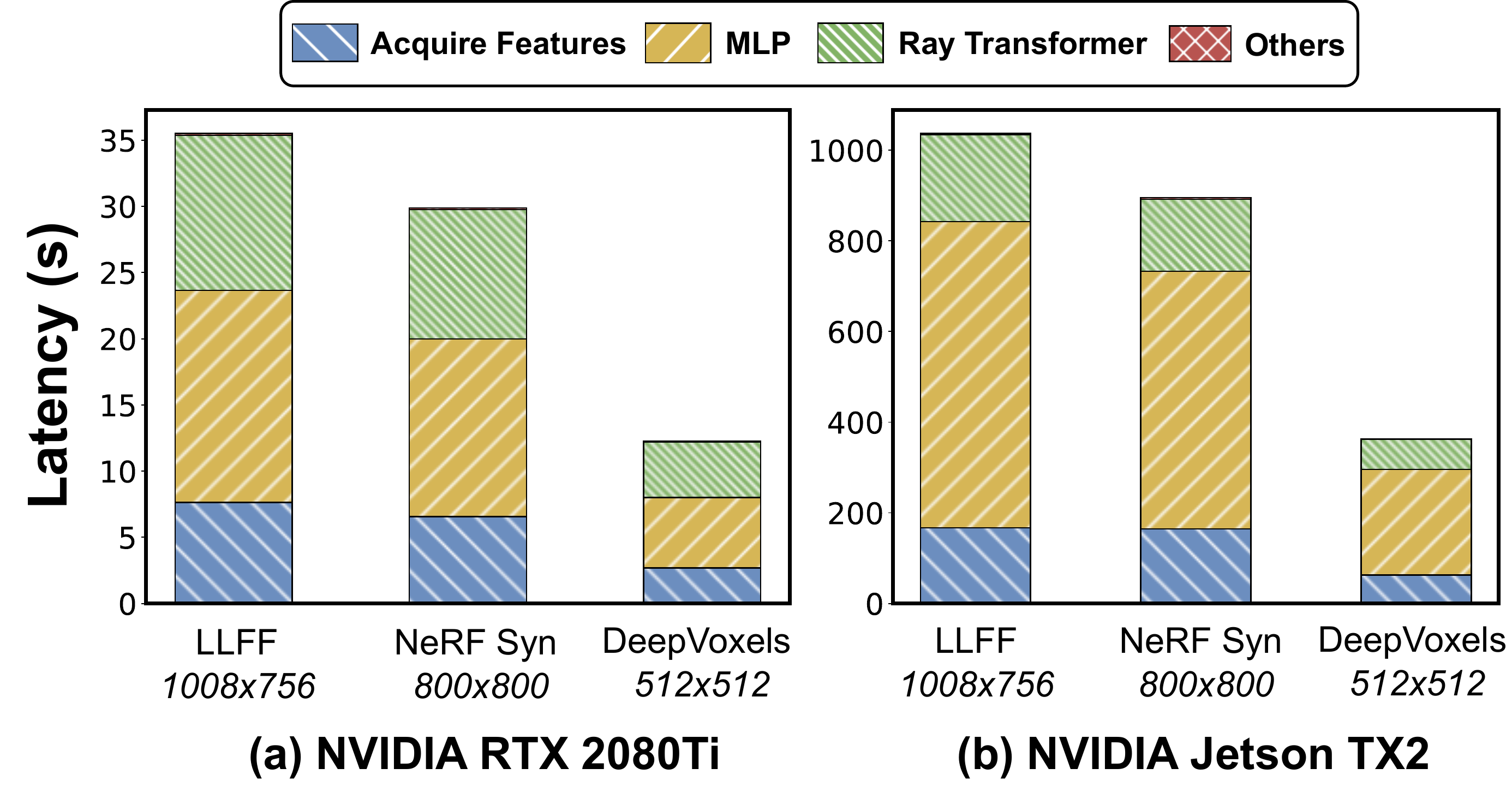}
\vspace{-1em}
\caption{Profile our generalizable NeRF model on two GPU devices across three datasets with different resolutions.}
\label{fig:profile}
\vspace{-1.5em}
\end{figure}

\textbf{Observations.}
As shown in Fig.~\ref{fig:profile}, we can observe that (1) the real-time requirement cannot be satisfied on both devices, e.g., RTX 2080Ti can only achieve a $\leqslant$0.249 FPS; (2) even if more computing resources are available to reduce the DNN inference time, the significant overhead for acquiring scene features will still prohibit the real-time execution; and (3) the ray transformer counts for 44.1\% of the total DNN inference time on RTX 2080Ti while its floating-point operations (FLOPs) counts for only 13.8\% of total DNN FLOPs on LLFF~\cite{mildenhall2019local}, indicating that the attention operations may not be well accelerated by RTX 2080Ti.

\subsection{Identified Opportunities for Acceleration}
\label{sec:opportunites}

\textbf{Unique sparsity opportunities in 3D scenes.}
The sources of sparsity stem from the varying contributions of sampled points in different regions to the rendered pixels. Specifically, camera rays emitted through regions, (1) with low particle density or (2) with low accumulated transmittance due to occlusion, require fewer sampled points without sacrificing rendering quality.
We hypothesize that properly leveraging these sparsity opportunities in 3D scenes can reduce the total number of points that are needed to be sampled while maintaining the rendering quality. In this way, the data movement and computational cost of Steps \ballnumber{2}-\ballnumber{4} corresponding points that do not have to be sampled can be skipped, thus improving the achievable frame-per-second (FPS).

\textbf{Potential scene feature reuses across rays and points.} 
The sampled 3D points from different rays may be projected to the same 2D points on the image planes of the source views. Therefore, there exist opportunities to reuse scene features across those points. As such, we hypothesize that properly designing the algorithm-to-hardware dataflow to map the rays/points of which the scene features can be reused can reduce the overall data movement cost of generalizable NeRFs. However, since the position and view direction of a user can arbitrarily change during run-time, the geometric relationship among different rays emitted from a user’s camera center is unknown before NeRF deployment, i.e., whether different rays/points can reuse the same scene features is uncertain. It is thus highly desirable to develop techniques that can efficiently derive the geometric relationship among rays given the position and view direction of a user's camera and then derive the point-to-hardware mapping schemes accordingly to maximize scene feature reuses.

\begin{figure*}[!t]
\centering
\vspace{-0.5em}
\includegraphics[width=0.95\linewidth]{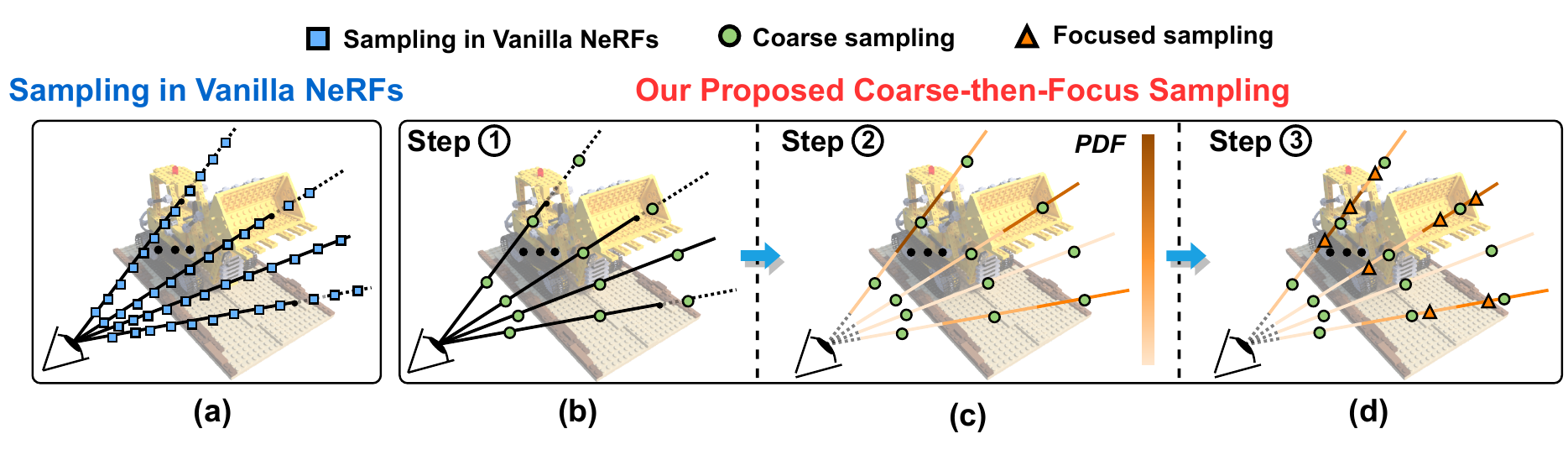}
\vspace{-1.5em}
\caption{
Illustrating (left) the sampling process in vanilla NeRFs and (right) our coarse-then-focus sampling strategy. 
}
\label{fig:coarse_then_focus}
\vspace{-1em}
\end{figure*}

\vspace{-0.5em}
\section{\METHOD{}: Algorithm}
\label{sec:alg}

\subsection{Algorithm Overview}
\label{sec:alg_overview}

Our \METHOD{}’s algorithm aims to (1) leverage the sparsity opportunities in target 3D scenes as analyzed in Sec.~\ref{sec:opportunites} to reduce the total number of required sampling points and thus corresponding data movement and computational costs while maintaining the rendering quality, and (2) reduce the heterogeneity of generalizable NeRF workloads caused by diverse computation patterns between the ray transformer $T$ and the MLP $f$ modules, thus enhancing the ease of acceleration. 
Our \METHOD{} fulfills these two objectives via developing a coarse-then-focus sampling strategy and an MLP-based Ray-Mixer module as an efficient alternative of the ray transformer as elaborated in Sec.~\ref{sec:alg_sampling} and Sec.~\ref{sec:alg_raymixer}, respectively.

\vspace{-0.5em}
\subsection{The Proposed Coarse-then-Focus Sampling}
\label{sec:alg_sampling}

\vspace{-0.2em}
\textbf{The overall pipeline.}
Fig.~\ref{fig:coarse_then_focus} shows our coarse-then-focus sampling strategy featuring three steps: Step \ding{172} performs lightweight coarse sampling to acquire the density distribution of the target 3D scene; Step \ding{173} identifies and filters the empty/occluded regions based on the estimated density distribution from the previous step and derives the sampling probability density function (PDF); Step \ding{174} conducts focused sampling based on the sampling PDF above, where the sampled points are non-uniformly and sparsely distributed across different rays. Finally, the sampled points are processed, following the vanilla generalizable NeRF pipeline (i.e., Steps \ballnumber{2}-\ballnumber{5} in Sec.~\ref{sec:workload}) to obtain the rendered pixels. 
Next, we elaborate Steps \ding{172}-\ding{174} below.

Step \ding{172}: Lightweight coarse sampling.
As the goal of this step is only to estimate the density distribution for identifying empty or occluded regions and deriving the sampling PDF in the next step, we find that its complexity can be aggressively trimmed down without hurting the final rendering quality.
Therefore, we adopt a lightweight design to implement coarse sampling by conditioning the NeRF process on only a limited number of views (denoted as $S_{c}$) that are the closest to the user's view direction and also sampling fewer points ($N_{c}$) along each ray as shown in Fig.~\ref{fig:coarse_then_focus} (b).

Step \ding{173}: Empty/occluded region discovery and sampling PDF estimation. The regions that contribute less to the rendered pixels caused by low density $\sigma_k$ (i.e., empty regions) and/or low accumulated transmittance $T_k$ (i.e., occluded regions) can be identified based on the hitting probability $w_k=T_k(1-\exp(-\sigma_k (t_{k+1}-t_{k})))$ according to Eq.~\ref{eq:nerf_render}, i.e., regions containing sampled points with low hitting probability $w_k < \tau$, obtained from the previous coarse sampling step, are considered to be unimportant for the rendered pixels, where $\tau$ is a predefined threshold. 
To leverage this opportunity toward sparse and thus more efficient sampling, we filter out these regions and assign more sampled points to important regions that contribute more to the rendered pixels in the subsequent focused sampling. To achieve this, we define sampled points with $w_k \geq \tau$ obtained from the coarse sampling step as critical points, and the probability $P(j)$ of sampling from the $j$-th ray is proportional to the number of critical points $N_j^{cr}$ on that ray, i.e., $P(j) = N_j^{cr}/ \sum_j N_j^{cr}$. As such, the  PDF $P(k,j)$ of sampling the $k$-th point on the $j$-th ray in the subsequent focused sampling is set as $P(k,j)=P(k|j) \cdot P(j)$ where $P(k|j) = w_k^j / \sum_k w_k^j $ and $w_k^j$ is the hitting probability of the $k$-th point on the $j$-th ray, following~\cite{mildenhall2020nerf}. This strategy produces a piecewise-constant PDF along each ray as shown in Fig.~\ref{fig:coarse_then_focus} (c).

Step \ding{174}: Sparse focused sampling.
We further sample another set of $H \times W \times N_f$ points in the 3D space based on the calculated sampling PDF above, where $N_f$ is the average number of sampled points per ray and $H$/$W$ denotes the height/width of the rendered image, respectively. Different from uniform sampling, the resulting sampled points are non-uniformly distributed across the rays, where fewer points are allocated on empty/occluded regions, thus allowing higher sampling sparsity while maintaining the rendering quality.

During the training process, the number of sampled points on each ray is required to be constant for facilitating the commonly used batch training of  NeRFs. To satisfy this constraint, 
we first perform focused sampling based on the sampling PDF in Step \ding{173} and then pad the sampled points along all rays to $N_{\max}$, a predefined maximal number of sampled points per camera ray. Note that the padded ones do not contribute to the volume rendering in Eq.~\ref{eq:nerf_render} and also do not participate in the rendering process at run-time.

\begin{figure*}[!t]
\centering
\vspace{-0.5em}
\includegraphics[width=0.95\linewidth]{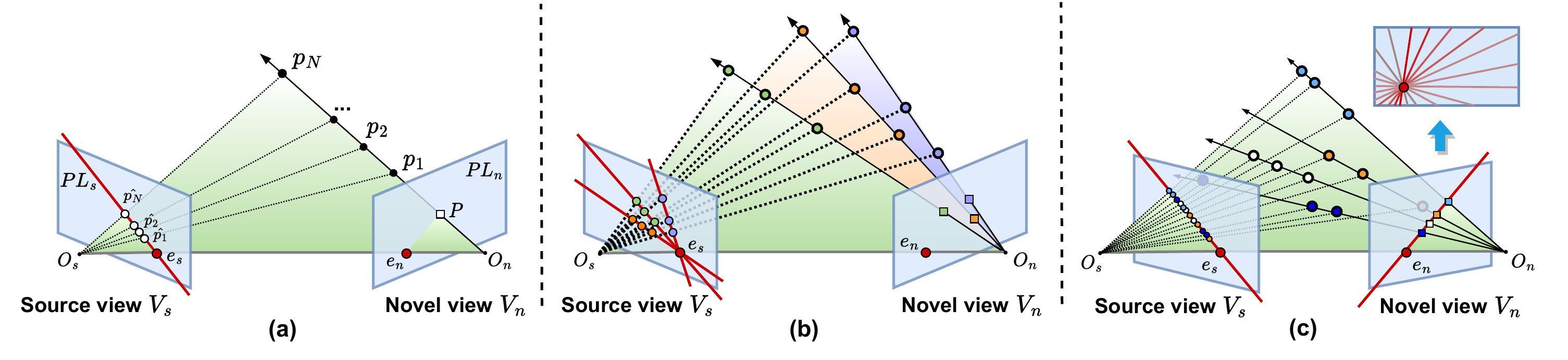}
\vspace{-1.5em}
\caption{Visualize the epipolar geometric relationship among (a) the sampled 3D points and their projections on the source view for one pixel/ray, (b) those for multiple pixels/rays, and (c) the rays with corresponding pixels located on the same line that passes through the epipole $e_n$.
These geometric relationships are deductions of the epipolar geometric analysis in \cite{hartley2003multiple}.
}
\label{fig:epipolar_geo}
\vspace{-1em}
\end{figure*}

\textbf{Differences from NeRF's hierarchical volume sampling.} 
Our coarse-then-focus sampling is built on top of the hierarchical volume sampling in vanilla NeRF~\cite{mildenhall2020nerf}, which aims to refine the details by rendering at two levels of granularity while sampling the same numbers of points across different rays. In contrast, our coarse-then-focus sampling has two major differences: (1) it leverages the sparsity in 3D scenes to trim down the complexity of the ray marching process, which is implemented by adopting a super lightweight coarse sampling only to predict the sampling PDF of the target 3D space without reconstructing the RGB value to boost efficiency; and (2) our focused sampling results in different numbers of sampled points across different rays depending on the estimated sampling PDF, enabling more sampled points on important regions.

\vspace{-1em}
\subsection{The Proposed Ray-Mixer Module}
\label{sec:alg_raymixer}
Considering that the ray transformer $T$ introduced in Sec.~\ref{sec:workload} is executed once per ray whereas the MLP model $f$ is executed for inferring every sampled point along the ray, designing a customized module to accelerate the ray transformer's attention operations is inefficient in terms of area and can cause hardware under-utilization. As such, it is highly desirable to unify the computation patterns of the MLP and ray transformer modules for boosted efficiency.

Inspired by~\cite{tolstikhin2021mlp}, we achieve the aforementioned goal by developing a module dubbed Ray-Mixer to fuse the density features $\{f_k^{\sigma}\}_{k=1}^N$ of all points along the same ray, which is to replace the attention operations commonly used in generalizable NeRFs~\cite{wang2021ibrnet,chen2021mvsnerf,reizenstein2021common,liu2022neural}. This module is implemented using three fully connected (FC) layers, eliminating the attention operations and thus reducing the heterogeneity in the required workload. In particular, for $f^{\sigma} \in \mathbb{R}^{N \times D}$, where $N$ and $D$ are the numbers of points along the ray and the feature dimension, respectively, our Ray-Mixer adopts the first FC layer along the point dimension to fuse the information across all sampled points on the same ray and then adopts another FC layer along the feature dimension for independently processing each sampled point to deliver their density prediction. We formulate the execution process of our Ray-Mixer as follows:

\begin{align}
\label{eq:ray_mixer}
F^{\sigma}_{*, i} &= f^{\sigma}_{*, i} + \phi\bigl( \mathbf{W}_1\, f^{\sigma}_{*, i} \bigr),\,\,\text{for }i=1\ldots D\\ 
\sigma_{j} &= \mathbf{W}_3 (F^{\sigma}_{j,*} + \phi\bigl( \mathbf{W}_2\, F^{\sigma}_{j,*} \bigr)),\,\,\text{for }j=1\ldots N 
\end{align}

\noindent where $\mathbf{W}_1$ and $\mathbf{W}_2$ are the weights of the two aforementioned FC layers,  $\mathbf{W}_3$ is the weight of a projection layer from density features to estimated densities, and $\phi$ is an activation function.

\section{\METHOD{}: Hardware}
\label{sec:hardware}

Our \METHOD{} accelerator leverages the deductions of epipolar geometry~\cite{hartley2003multiple} to analyze and accelerate the target workloads to maximize the data reuses among rays. 
To present our accelerator, we first introduce the basics and deductions of epipolar geometry referring to \cite{hartley2003multiple} and then identify corresponding scene feature reuse opportunities in Sec.~\ref{sec:epipolar}, leveraging which we develop our accelerator with optimized dataflow, feature storage format, and micro-architecture in Sec.~\ref{sec:dataflow}-Sec.~\ref{sec:architecture}, respectively.

\vspace{-1em}
\subsection{Epipolar Geometric Analysis}
\label{sec:epipolar}

As the ray marching process in generalizable NeRFs is performed in a 3D space, the memory access patterns of different rays, which are emitted from the novel views determined by the user at run-time, are complex to be analyzed. As such, it is highly desired to map such workloads into a 2D space for better analyzing the ray behaviors. 

\textbf{Epipolar geometry in generalizable NeRFs.}
Epipolar geometry~\cite{hartley2003multiple} is used to depict the geometric relations among the pixels (as well as the corresponding rays and sampled 3D points) observed from different viewpoints of a 3D scene, i.e., the novel views and source views in our case. Specifically, it infers the projections of sampled 3D points on the rays emitted from novel views onto the 2D image planes of corresponding source views and thus can be leveraged to analyze the scene feature access patterns.

\textbf{Basics of epipolar geometry.}
Fig.~\ref{fig:epipolar_geo} (a) visualizes the geometric relationship among the sampled 3D points along two specific rays emitted from the novel view $V_n$ and their projections on the 2D image planes on one of the source views $V_s$, where ${O_n}$ and ${O_s}$ are the camera centers for $V_n$ and $V_s$, respectively, and ${PL_n}$/${PL_s}$ are the corresponding novel/source image planes.
A more general case of the geometry among multiple rays is visualized in Fig.~\ref{fig:epipolar_geo} (b).
Formally, 
the triangle plane defined by the connection ${O_n O_s}$ and the ray ${O_n p_N}$ is called epipolar plane (highlighted in green)
and the intersections between line $O_n O_s$ and the two image planes $e_n$ and $e_s$ are called epipoles.

\textbf{Our leveraged property.}
We intend to analyze the access patterns and reuse opportunities of source features by leveraging the following property as analyzed in~\cite{hartley2003multiple}:

\textit{Property-1}: For sampled 3D points along the same ray, their projections on the source image plane ${PL_s}$ are on the same line with its corresponding epipole $e_s$, which is dubbed an epipolar line.

In particular, to render a pixel $P$ on novel images, $N$ points $\{p_i\}_{i=1}^N$ are sampled from the ray $OP$, which are projected to $\{\hat{p_i}\}_{i=1}^N$ on the source image plane ${PL_s}$ as introduced in Step \ballnumber{2} in Sec.~\ref{sec:workload}. 
\textit{Property-1} indicates that $\{\hat{p_i}\}_{i=1}^N$ are on the same epipolar line $e_s \hat{p_N}$, along which the scene features should be acquired to render the same pixel $P$. We thus leverage this to analyze the memory access patterns of the feature acquisition step of generalizable NeRFs, as elaborated below.

\textbf{Projection locality for scene feature reuse.} 
\textit{Property-1} reveals the memory access patterns of generalizable NeRFs via mapping the 3D ray marching process onto 2D image planes, which inspires us that the sampled 3D points that can be projected to locally close regions on the 2D image planes can enjoy the opportunities of scene feature reuses. This observation holds for both 3D points along the same ray and those across different rays, considering the 3D points on different rays are likely to be projected to the same or close regions on the 2D image planes, which is determined by the geometric relationship among rays. This calls for both customized dataflows that can leverage the projection locality mentioned above to maximize scene feature reuses and customized micro-architecture that can derive the geometric relationship among rays at run-time.

To leverage projection locality during point-to-hardware mappings, we start from a special but highly desired case when only one source view is available in Sec.~\ref{sec:case_study} and then extend to cases with multiple source views in Sec.~\ref{sec:dataflow}.

\begin{figure*}[!t]
\centering
\vspace{-0.5em}
\includegraphics[width=0.95\linewidth]{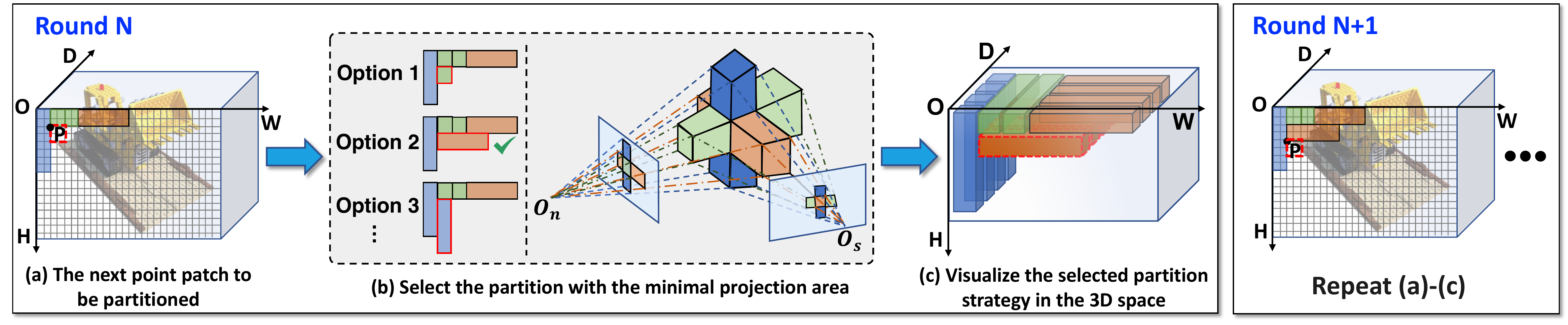}
\vspace{-1em}
\caption{Visualize our greedy 3D-point-patch partition that divides sampled 3D points into patches to maximize data reuses.}
\label{fig:dataflow}
\vspace{-1em}
\end{figure*}

\vspace{-0.3em}
\subsection{A Case Study with One Single Source View}
\label{sec:case_study}
Generalizing NeRFs to new scenes with only one source view available is an extreme but highly desired setting, which eases users' efforts for manually capturing the new scene of interest and thus has gained increasing attention~\cite{xu2022sinnerf, yu2021pixelnerf}. 
We customize the dataflow under a single source view based on the following property that we infer from \textit{Property-1}:

\textit{Property-2}: The rendered pixels located on the same line that passes through the epipole $e_n$ on the plane ${PL_n}$ share the same epipolar line on the source image plane ${PL_s}$.
More specifically, as shown in Fig.~\ref{fig:epipolar_geo} (c), the rays emitted from the pixels located on the same line that passes through the epipole $e_n$ are on the same epipolar plane and thus they share the same epipolar line, which is geometrically the intersection between the epipolar plane and the source image plane ${PL_s}$.

\textbf{Customized dataflow under single source view.} 
\textit{Property-2} indicates that the rays corresponding to the pixels located on the same line that passes through the epipole $e_n$ can share the scene feature reuse opportunities thanks to their shared epipolar line. To leverage this, a simple but effective solution is to prioritize the rendering on pixels that share the same epipolar line via simultaneously mapping them to the hardware. In particular, as shown in the top-right part of Fig.~\ref{fig:epipolar_geo} (c), we divide the pixels (and the corresponding rays) into groups via drawing lines that pass through the epipole $e_n$, where each line corresponds to one ray group. The rays in the same ray group are emitted and processed simultaneously, following Steps \ballnumber{1}-\ballnumber{5} in Sec.~\ref{sec:workload}, and thus the scene features of their sampled 3D points can be reused via being fetched and buffered only once by our micro-architecture introduced in Sec.~\ref{sec:architecture}.

\textbf{Challenges for extending to multiple source views.} 
However, the aforementioned solution is not directly applicable to new scenes with multiple source views. This is because the epipole $e_n$ is defined over a pair of views, i.e., the epipoles between the novel view and different source views are different, and thus the rays that share the same epipolar line on one specific source view may not necessarily share the epipolar line on another source view, thus diminishing the effectiveness of the aforementioned dataflow.
Therefore, a more principled point-to-hardware mapping strategy is required for handling new scenes with multiple source views.

\vspace{-0.8em}
\subsection{The Proposed Point-to-Hardware Mappings}
\label{sec:dataflow}

To make use of the scene feature reuse opportunities for new scenes with varied numbers of source views, we identify the following deduction from epipolar geometry~\cite{hartley2003multiple}, which depicts the spatial locality of epipolar geometry:

\textit{Property-3}: The sampled 3D points that are close in 3D locations will share close epipolar lines on the source views no matter being observed from any novel view, implying that the source features on their epipolar lines can be simultaneously acquired and processed.

\textbf{Our proposed dataflow.}
\textit{Property-3} motivates us to simultaneously map the rays that share the closer depth and view directions onto the hardware, where the achievable acceleration efficiency is determined by how to divide the 3D points into patches that are simultaneously processed.
To achieve this, we describe the target workload as a 3D cube that covers the information of both view directions and depths. In particular, for rendering a 2D image with a resolution of $H \times W$ and $N$, $N$ sampled points along each ray lie in different depths of the corresponding ray, and the depth range between the near plane to the far plane is denoted as $D$, where the aforementioned 3D cube features a shape of $H \times W \times D$. For scheduling the point-to-hardware mapping, our proposed dataflow slices a point patch $\delta h \times \delta w \times \delta d$ from the 3D cube, which is prefetched and processed at one time to exploit the scene feature reuse.

As analyzed in Sec.~\ref{sec:epipolar}, the scene feature reuse opportunities among points are determined by their geometric relationship, which keeps changing along with the movement of novel view directions from the users' perspective at run-time. Therefore, it is highly desirable to efficiently derive the optimal 3D-point-patch partition strategy, i.e., the patch shape $\delta h \times \delta w \times \delta d$ per processing, at run-time based on the epipolar geometric analysis to fully unleash the potential of our dataflow.

\textbf{Proposed greedy 3D-point-patch partition.}
We propose a greedy 3D-point-patch partition algorithm to divide the 3D workload cube $H \times W \times D$ into point patches at run-time. 
The rationale of our strategy is to iteratively select the partition strategy for each local region that contains the same number of 3D sampled points while greedily minimizing the required memory accesses of scene features at each iteration. The reason that we adopt a greedy scheme is to ease the corresponding hardware implementation for reduced run-time overhead.

As shown in Fig.~\ref{fig:dataflow}, our algorithm gradually partitions the 3D workload cube into patches from the top-left in the near plane, i.e., $(h,w,d)=(0,0,0)$, to the bottom-right in the far plane, i.e., $(h,w,d)=(H,W,D)$. To decide the shape of each point patch, we greedily search for the optimal patch shape, which aims to maximize the scene feature reuse opportunities, among $M$ predefined patch shape candidates $\{\delta h_i, \delta w_i, \delta d_i\}_{i=1}^M$ shown in the left part of Fig.~\ref{fig:dataflow} (b). 
More specifically, as visualized in the right part of Fig.~\ref{fig:dataflow} (b), a point patch candidate essentially constructs a frustum in the 3D world coordinate when being transformed from the $(h,w,d)$ space. We leverage the epipolar geometry to project each frustum onto the 2D image planes of corresponding 
source views, and use the total covered area of the projection, which is a 2D tetragon that covers the epipolar lines of all rays within the frustum, to estimate the required memory access for processing this point patch candidate. After selecting the optimal patch shape $\{\delta h_{opt}, \delta w_{opt}, \delta d_{opt}\}$ that minimizes the memory access at each iteration, we assign the points that fall into the corresponding frustum in the 3D space to the same patch, which is pushed into a patch queue for sequential processing. We iteratively perform the partition until all sampled 3D points are assigned to a patch.

Note that we enforce two constraints during this scheduling process: (1) we enforce the patches located at the same height $h$ and width $w$ but different depth $d$, which corresponds to the same set of pixels, to share the same patch partition strategy to ease the hardware logic when accumulating their predicted colors in Step \ballnumber{5} in Sec.~\ref{sec:workload}, and (2) we guarantee that the total read size during prefetching, which is determined by the patch shape, does not exceed the maximum prefetch buffer size.

\vspace{-0.7em}
\subsection{Optimize the Feature Storage Strategy}
\label{sec:storage}

\begin{figure}[!t]
\centering
\includegraphics[width=0.99\linewidth]{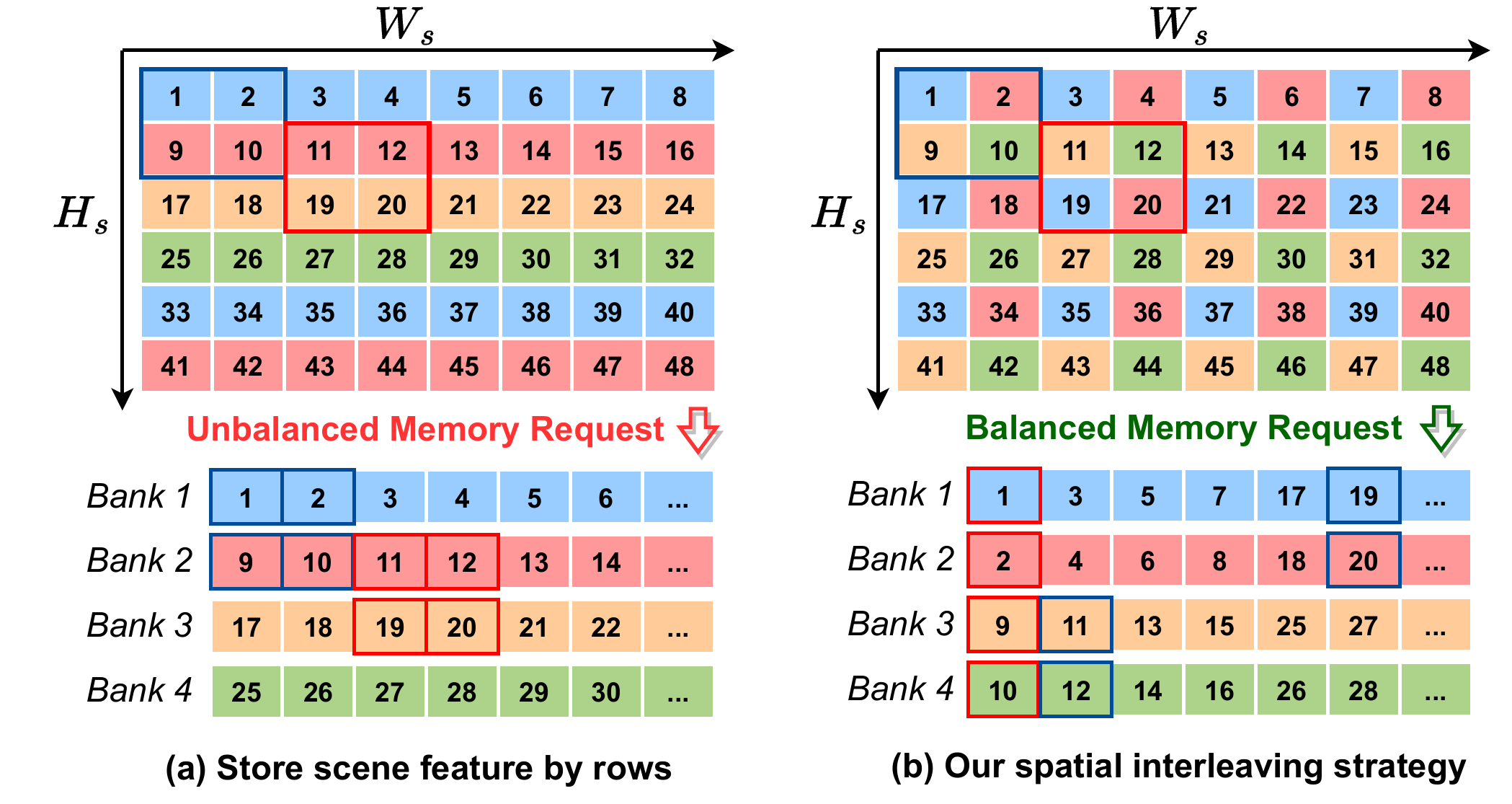}
\vspace{-1em}
\caption{Visualize different strategies for storing features.}
\label{fig:storage}
\vspace{-2em}
\end{figure}

Considering the scene features with a shape of $S \times H_s \times W_s \times C$, where $S$ is the number of source views, $H_s$/$W_s$ are the feature height/width, respectively, and $C$ is the feature dimension, are stored in DRAM and prefetched to a prefetch buffer for each point patch, an improper storage strategy, as shown in Fig.~\ref{fig:storage} (a), may result in memory bank conflicts when simultaneously querying the scene features stored on the same memory bank, resulting in reduced bandwidths. To this end, we optimize the storage strategy of scene features to balance the read/write volume between the memory banks.

\textbf{Our proposed strategy.}
Motivated by the projection locality introduced by our dataflow, which simultaneously accesses the scene features in a local region on the 2D image planes of the source views, it can be highly expected that the scene features in the 2D local region can be read out without bank conflicts. To achieve this, we propose a spatial interleaving storage format along the $H_s$ and $W_s$ dimensions as shown in Fig.~\ref{fig:storage} (b), where neighboring features on the image plane are stored in different memory banks so that the bank conflicts can be alleviated for prefetching each point patch from a multi-bank DRAM.

\subsection{Gen-NeRF's Micro-architecture}
\label{sec:architecture}

\begin{figure*}[!t]
\centering
\includegraphics[width=0.95\linewidth]{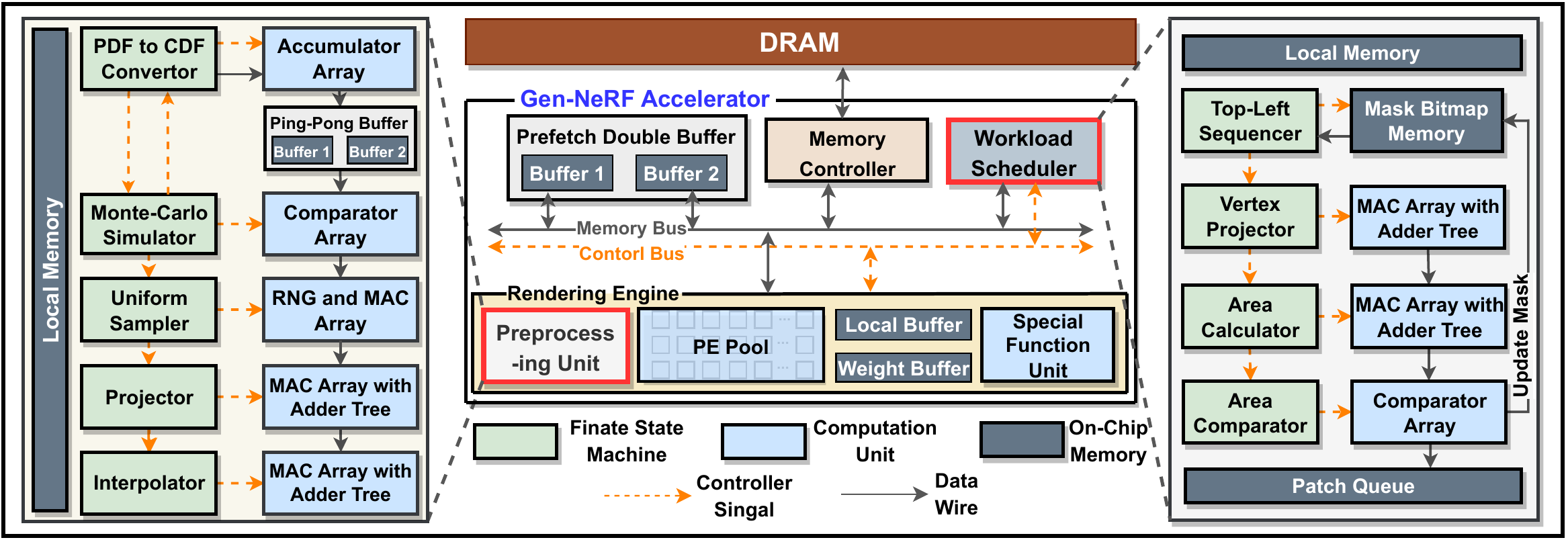}
\vspace{-1em}
\caption{An illustration of the micro-architecture in our \METHOD{} accelerator, where the block diagram is shown in the middle and two blocks are further visualized at the left and right parts, including the preprocessing unit and workload scheduler.}
\label{fig:architecture}
\vspace{-0.5em}
\end{figure*}

\textbf{Micro-architecture overview.}
Fig.~\ref{fig:architecture} shows our \METHOD{} accelerator's micro-architecture, integrating a memory controller to handle the communication with the off-chip DRAM, a prefetch double buffer to store scene features for data reuses, a workload scheduler to partition the workloads and support the dataflow introduced in Sec.~\ref{sec:dataflow}, and a rendering engine to perform the coarse/focused sampling and ray marching.

\textbf{Rendering on \METHOD{}'s micro-architecture}.
Here we describe how the rendering process is executed in our \METHOD{}'s micro-architecture (see Fig.~\ref{fig:architecture}). Given a sparse set of source views of a scene and a novel view from the user, the workload scheduler continuously performs the greedy 3D-point-patch partition in Sec.~\ref{sec:dataflow} to generate and enqueue point patches when the patch queue is not full, which are next processed by the prefetch buffer and the rendering engine.
The prefetch buffer is executed in a double-buffer manner to hide the off-chip communication latency, i.e., when one of the SRAMs in the prefetch buffer is used to provide scene features for the rendering engine, the other SRAM will be used to prefetch the next point patch from the patch queue.

\begin{figure}[th]
\centering
\vspace{-1em}
\includegraphics[width=\linewidth]{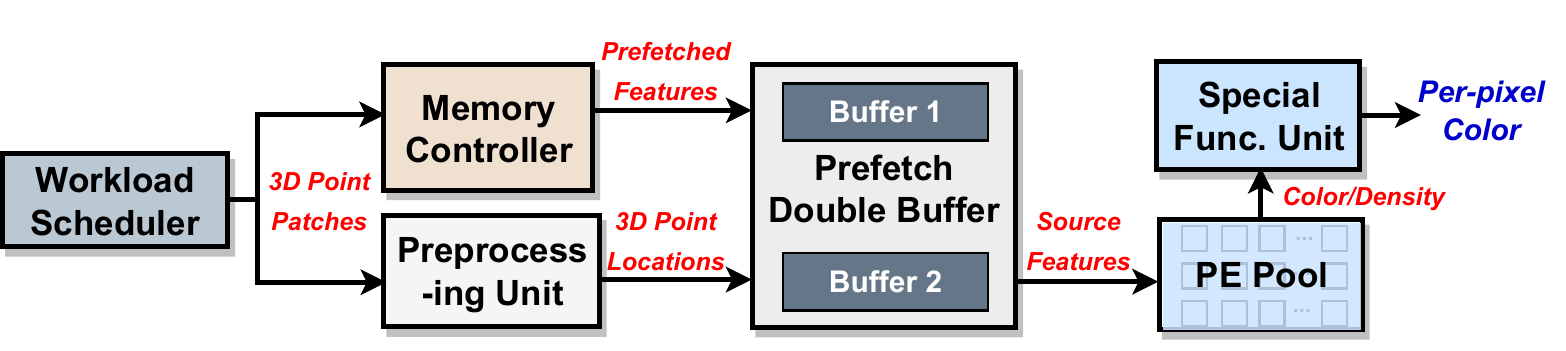}
\vspace{-2.2em}
\caption{Visualize the workflow of a single stage in our \METHOD{}'s rendering process.}
\label{fig:workflow}
\vspace{-2em}
\end{figure}

The rendering engine features two execution stages for supporting our coarse-then-focus sampling in Sec.~\ref{sec:alg_sampling}. In the first stage, the rendering engine performs the lightweight coarse sampling for the scheduled point patch; In the second stage, it performs sampling PDF estimation, focused sampling, and volume rendering.
Both stages can be further divided into Steps \ballnumber{1}-\ballnumber{5} as mentioned in Sec.~\ref{sec:workload}, where Steps \ballnumber{1}-\ballnumber{4} are carried out in a pipelined manner and the ray transformer in Step \ballnumber{4} is replaced with our Ray-Mixer. In particular, based on the partitioned point patches from the workload scheduler, the preprocessing unit samples 3D points and acquires their corresponding source features from the prefetch double buffer (i.e., Steps \ballnumber{1}-\ballnumber{2}). Next, the source features are fed to the PE pool for executing the MLP and our Ray-Mixer (i.e., Steps \ballnumber{3}-\ballnumber{4}). Finally, the predicted density and colors from Step \ballnumber{4} are written into the local memory of the rendering engine and are used to generate the final pixels in Step \ballnumber{5} after the pipeline is done for the patch. Fig.~\ref{fig:workflow} shows the corresponding workflow diagram of only one stage.

\vspace{0.3em}
\textbf{Design details of each featured component.}
The prefetch buffer is a pair of SRAMs, forming a double buffer to enable the buffer to read and write in parallel. Each SRAM is divided into multiple banks and we also leverage the spatial interleaving storage format introduced in Sec.~\ref{sec:storage} to store scene features for balancing the communication between SRAM banks in each prefetch buffer and thus improve the effective bandwidth.

The workload scheduler is composed of a top-left sequencer for indicating the next location to be processed based on a mask-bitmap memory, which is a bitmap to indicate whether a 3D point has been assigned to a patch during the iterative 3D-point-patch partition process, a vertex projector that projects 3D frustums in the 3D space into tetragons on source image planes, an area calculator and the corresponding comparator to derive the optimal patch shape, and a patch queue to store the partitioned point patches.

The rendering engine comprises a preprocessing unit, a local buffer, a weight buffer, a PE pool, and a special function unit. In particular, the preprocessing unit handles (1) the focused sampling from the estimated sampling PDF in Sec.~\ref{sec:alg_sampling} using inverse transform sampling in Monte-Carlo methods~\cite{sobol2018primer} and (2) the loading and preprocessing of scene features via first projecting the sampled points onto source image planes by the projector and then bilinearly interpolating the exact scene features among those of the closest four elements by the interpolator.
The PE pool is composed of multiple PEs, each of which is a systolic array, to execute the MLP and Ray-Mixer modules. The special function unit is composed of a PE line to calculate the exponential function and accumulate the colors of sampled points along the ray in Eq.~\ref{eq:nerf_render}.

\textbf{Comparisons with GPUs.}
If we compare our architecture with GPUs~\cite{cuda_guideline},
the roles of the PE pool and the prefetch double buffer in Fig.~\ref{fig:architecture} could emulate the thread blocks and L2-cache that can be accessed by different blocks, respectively. Our architecture differs in (1) the preprocessing unit that handles our proposed point sampling strategy introduced in Sec.~\ref{sec:alg_sampling}, and (2) the workload scheduler that implements our proposed 3D-point-patch partition to maximize source feature reuses.

\section{Experimental Results}
\label{sec:experiment}

\subsection{Experiment Setup}
\label{exp:setup}

\textbf{Datasests.} We use the training sets in~\cite{wang2021ibrnet}, which integrates
both synthetic data and real data from  Google Scanned Objects~\cite{google_scanned_objects}, RealEstate10K~\cite{zhou2018stereo}, the Spaces dataset~\cite{flynn2019deepview}, LLFF~\cite{mildenhall2019local}, and self-captured real scenes from handheld cellphones~\cite{wang2021ibrnet}.
For evaluation, we follow~\cite{mildenhall2020nerf,wang2021ibrnet} and use both synthetic objects and real scenes, including four Lambertian objects from DeepVoxels~\cite{sitzmann2019deepvoxels}, eight synthetic objects from~\cite{mildenhall2020nerf}, and 
eight complex real-world scenes captured with roughly forward-facing images from LLFF~\cite{mildenhall2019local}.

\textbf{Algorithm setup.}
We build our framework on top of IBRNet~\cite{wang2021ibrnet} as it is the most representative generalizable NeRF design with its pipeline inherited by follow-up generalizable NeRF variants. We train the models for 250K steps using an Adam optimizer and an initial learning rate of 5e-4 with exponential decay, following~\cite{wang2021ibrnet}. 
During the lightweight coarse sampling, we fix the number of source views as 4 and apply a channel scale of 0.25 to the coarse MLPs. The typical workload of rendering an 800$\times$800 image, which samples 64 points per ray on average during the focused sampling and is conditioned on 6 source views, involves 0.328 trillion FLOPs.

\textbf{Hardware setup.}
We implement our hardware modules (e.g., the rendering engine and workload scheduler in Fig.~\ref{fig:architecture}) in Verilog and use Cadence Genus to synthesize the gate-level design for estimating the chip area, timing, and power consumption information based on a commercial 28nm CMOS technology. In particular, the synthesized frequency is set to 1GHz. The rendering engine consists of a PE pool with 40 16*16 INT8 systolic arrays, a 256KB local buffer, and an 8KB weight buffer. Each of the prefetch buffers is a 256KB scratchpad memory. The detailed area and power of each hardware module are provided in Tab.~\ref{tab:area_power}. 

Due to the lack of an RTL model for DRAM, an end-to-end verilog-simulation is not feasible. As such, we build a cycle-accurate simulator to characterize the behaviors of our accelerator based on (1) the timing and power information derived from gate-level simulations and (2) a commonly-used tool for DRAM, Ramulator~\cite{kim2015ramulator}, to estimate the DRAM latency/energy. Specifically, we calculate the number of cycles of each hardware module and record all memory requests to the off-chip DRAM, i.e., an LPDDR4-2400 DRAM with 17.8 GB/s bandwidth that is commonly used for AR/VR devices~\cite{quest_pro}.

\subsection{Evaluating \METHOD{}'s Algorithm}
\label{sec:evaluate_alg}

\begin{figure}[ht]
\centering
\vspace{-1em}
\includegraphics[width=\linewidth]{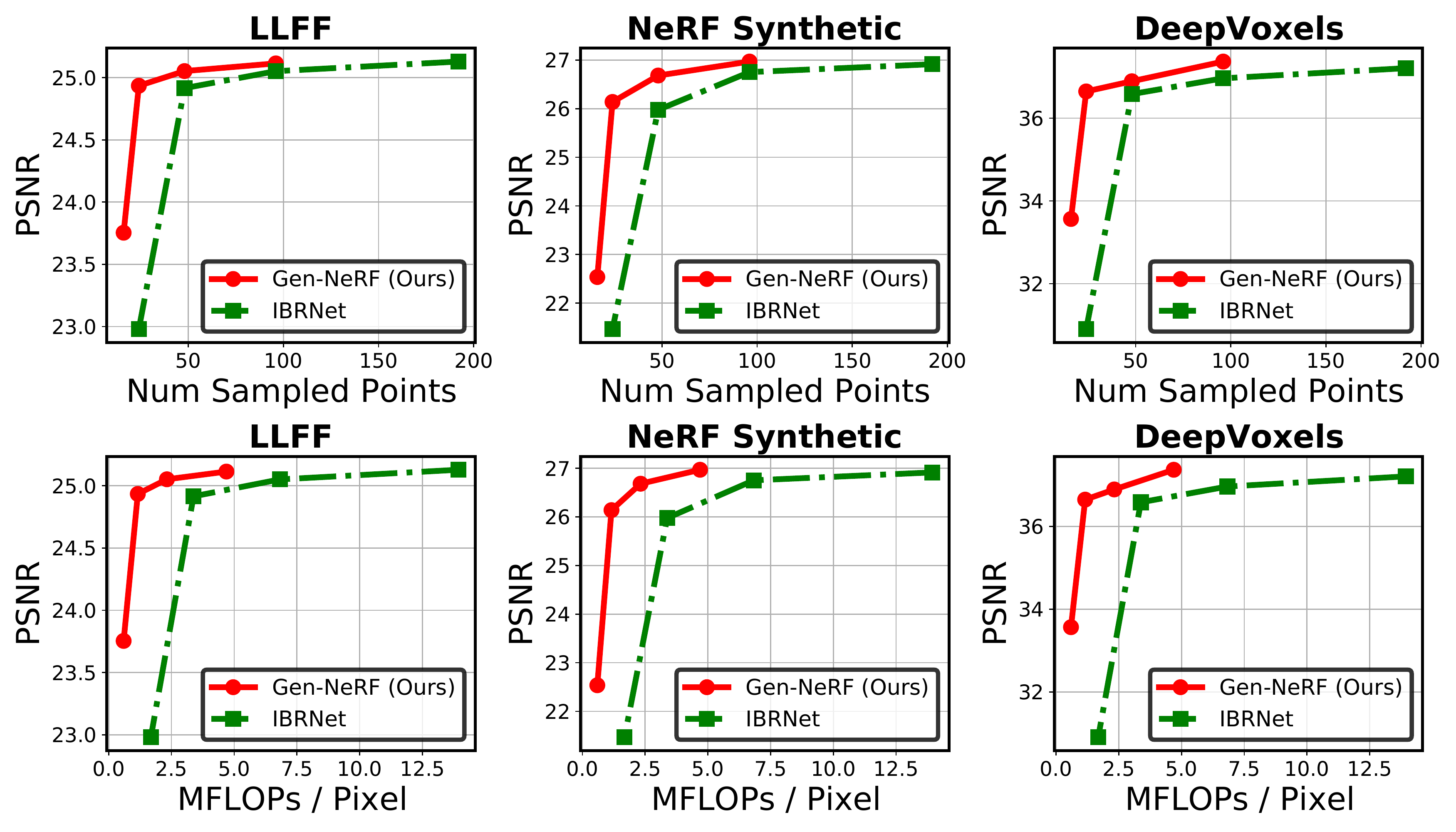}
\vspace{-2.5em}
\caption{Benchmark our \METHOD{} with IBRNet under different numbers of sampled points (top) and corresponding million FLOPs (MFLOPs) for rendering one pixel (bottom).}
\label{fig:num_sampled_points}
\vspace{-0.5em}
\end{figure}

\textbf{Effectiveness of our coarse-then-focus sampling strategy.} We first benchmark our coarse-then-focus sampling with the sampling strategy of IBRNet~\cite{wang2021ibrnet} under different numbers of sampled points. In particular, for \METHOD{}, we sample 8/8, 8/16, 16/32, and 32/64 points per camera ray on average during the coarse/focused sampling, respectively, and Fig.~\ref{fig:num_sampled_points} visualizes the achievable trade-off between PSNR (averaged over all scenes in each dataset) and the total numbers of sampled points/the corresponding FLOPs for rendering one pixel. We can observe that (1) our coarse-then-focus sampling consistently achieves a better PSNR under the same number of sampled points, e.g., a 4.67 higher PSNR with 24 sampled points on NeRF Synthetic, and (2) thanks to the superior efficiency of our lightweight coarse sampling, the required FLOPs for inferring the same number of points of our method is also reduced.

\begin{table}[t]
\centering
\caption{Area and power of \METHOD{}'s hardware modules.}
\vspace{-1em}
\resizebox{0.99\linewidth}{!}
{    
\begin{tabular}{c|cccc|c}
\toprule
\textbf{Module} & \textbf{\begin{tabular}[c]{@{}c@{}}Workload \\ Scheduler\end{tabular}} & \textbf{\begin{tabular}[c]{@{}c@{}}Preprocessing \\ Unit (PPU)\end{tabular}} & \textbf{\begin{tabular}[c]{@{}c@{}}Rendering Engine \\ (except PPU)\end{tabular}} & \textbf{\begin{tabular}[c]{@{}c@{}}Prefetch\\ Buffer\end{tabular}} & \textbf{Total} \\ \midrule
Area (mm$^2$) & 0.24 & 1.24 & 14.98 & 1.34 & 17.80 \\
Power (mW) & 156.2 & 696.0 & 8359.2 & 473.6 & 9685.0 \\ \bottomrule
\end{tabular}
}
\label{tab:area_power}
\vspace{-1em}
\end{table}

\textbf{Efficacy of our Ray-Mixer.} 
 We benchmark the following three IBRNet variants: (a) vanilla IBRNet with the ray transformer, (b) IBRNet with the ray transformer removed, and (c) IBRNet integrating our Ray-Mixer.
As shown in rows 2-4 of Tab.~\ref{tab:ablation}, we can observe that (1) removing the ray transformer leads to a significant PSNR drop due to erroneous density estimation~\cite{wang2021ibrnet}, and (2) integrating Ray-Mixer results in considerably better reconstruction accuracy over that w/o Ray-Mixer, e.g., a 3.37 PSNR improvement on average across four scenes from LLFF. In addition, our Ray-Mixer can achieve comparable rendering quality as the ray transformer while significantly reducing workload heterogeneity since hardware modules dedicated to attention operations are no more required.

\begin{table}[t]
\centering
\caption{Impact of each component on the rendering quality (PSNR$\uparrow$/LPIPS$\downarrow$) and efficiency (MFLOPs/pixel).}
\vspace{-1em}
\resizebox{\linewidth}{!}
{    
\begin{tabular}{lccccc}
\toprule
\textbf{Method} & \begin{tabular}[c]{@{}c@{}}\textbf{MFLOPs} \\ \textbf{/ pixel}\end{tabular} & \textbf{fern} & \textbf{fortress} & \textbf{horns} & \textbf{trex} \\ \midrule
vanilla IBRNet & 13.94 & \textbf{23.837/0.246} & 30.003/\textbf{0.153} & 26.477/\textbf{0.177} & \textbf{24.574/0.230} \\ 
- ray transformer & 13.25 & 19.380/0.406 & 25.927/0.282 & 23.244/0.276 & 21.970/0.310 \\ \midrule \midrule
+ \textit{Ray-Mixer} & 13.88 & 23.716/0.247 & 30.028/0.153 & 26.427/0.177 & 23.810/0.236 \\
\begin{tabular}[c]{@{}c@{}}+ \textit{Coarse-then-Focus} \\ \textit{Sampling (16/48)}\end{tabular} & 4.27 & 23.657/0.252 & \textbf{30.087}/0.154 & \textbf{26.512}/0.181 & 24.158/0.239 \\ 
+ \textit{channel pruning} &  &  &  &  &  \\
\,\,\,\,\,\,\,  $\cdot$ 10 source views & 0.80 & 23.258/0.266 & 29.418/0.163 & 25.723/0.198 & 23.733/0.251 \\
\,\,\,\,\,\,\,  $\cdot$ 6 source views & 0.51 & 22.554/0.284 & 28.904/0.178 & 25.168/0.212 & 23.327/0.261 \\
\,\,\,\,\,\,\,  $\cdot$ 4 source views & 0.37 & 22.226/0.302 & 27.879/0.193 & 24.508/0.227 & 22.694/0.279 \\ \bottomrule
\end{tabular}
}
\label{tab:ablation}
\vspace{-1em}
\end{table}

\textbf{Impact of each component on the rendering quality and efficiency trade-off.}
We demonstrate the impact of each of our techniques on both rendering quality and efficiency in Tab.~\ref{tab:ablation}, providing a detailed breakdown of their respective contributions on top of four scenes from LLFF. In addition to our coarse-then-focus sampling and Ray-Mixer, we also reduce the redundancy in the model structure via channel pruning with a sparsity of 75\%, ensuring a comparable rendering quality with a $<$0.5 PSNR reduction on average on LLFF, to achieve a better PSNR-efficiency trade-off to better satisfy real-world application requirements. We can observe that (1) our coarse-then-focus sampling and Ray-Mixer can reduce the required FLOPs by 3.26$\times$ while achieving a comparable or even slightly higher PSNR as compared to vanilla IBRNet, and (2) introducing channel pruning could result in a $>$5$\times$ extra FLOPs reduction and the delivered model using 6 source views can reduce the required FLOPs by 27.3$\times$ while maintaining a $<$1.3 PSNR drop.

\textbf{Benchmark under a per-scene finetuning setting.} Considering that a per-scene finetuning process on top of pretrained generalizable NeRFs is found to enhance the reconstruction accuracy on a specific scene~\cite{wang2021ibrnet, chen2021mvsnerf, liu2022neural}, we further finetune \METHOD{}'s delivered models in Tab.~\ref{tab:ablation} and benchmark with finetuned IBRNet on top of four scenes from LLFF. As shown in Tab.~\ref{tab:per_scene}, our \METHOD{} can significantly trim down the complexity of IBRNet by $>$17$\times$ while maintaining a comparable PSNR (-0.38$\sim$-0.90).

\begin{table}[t]
\centering
\caption{Benchmark the rendering quality (PSNR$\uparrow$/LPIPS$\downarrow$) and efficiency (MFLOPs/pixel) using per-scene finetuning.}
\vspace{-1em}
\resizebox{\linewidth}{!}
{    
\begin{tabular}{ccccccc}
\toprule
\textbf{\begin{tabular}[c]{@{}c@{}}\# Source \\ Views\end{tabular}} & \textbf{Method} & \textbf{\begin{tabular}[c]{@{}c@{}}MFLOPs\\ / pixel\end{tabular}} & \textbf{fern} & \textbf{fortress} & \textbf{horns} & \textbf{trex} \\ \midrule
\multirow{2}{*}{4} & IBRNet & 6.31 & 23.759/0.247 & 29.893/0.165 & 27.149/0.162 & 25.293/0.217 \\
 & Gen-NeRF & 0.368 & 23.380/0.267 & 29.450/0.176 & 26.249/0.191 & 24.547/0.235 \\ \midrule \midrule
\multirow{2}{*}{10} & IBRNet & 13.94 & 24.890/0.210 & 31.237/0.139 & 28.471/0.141 & 26.644/0.199 \\
 & Gen-NeRF & 0.803 & 24.264/0.235 & 30.551/0.149 & 27.565/0.166 & 25.742/0.218 \\ \bottomrule
\end{tabular}
}
\label{tab:per_scene}
\vspace{-1.5em}
\end{table}

\subsection{Evaluating Gen-NeRF's Accelerator}
\label{sec:evaluate_hw}

We benchmark our \METHOD{}'s accelerator with both commercial GPUs and SOTA NeRF accelerators. We adopt 64 sampled points per ray on average and 6 source views if not specifically stated.

\begin{figure}[h]
\centering
\includegraphics[width=0.95\linewidth]{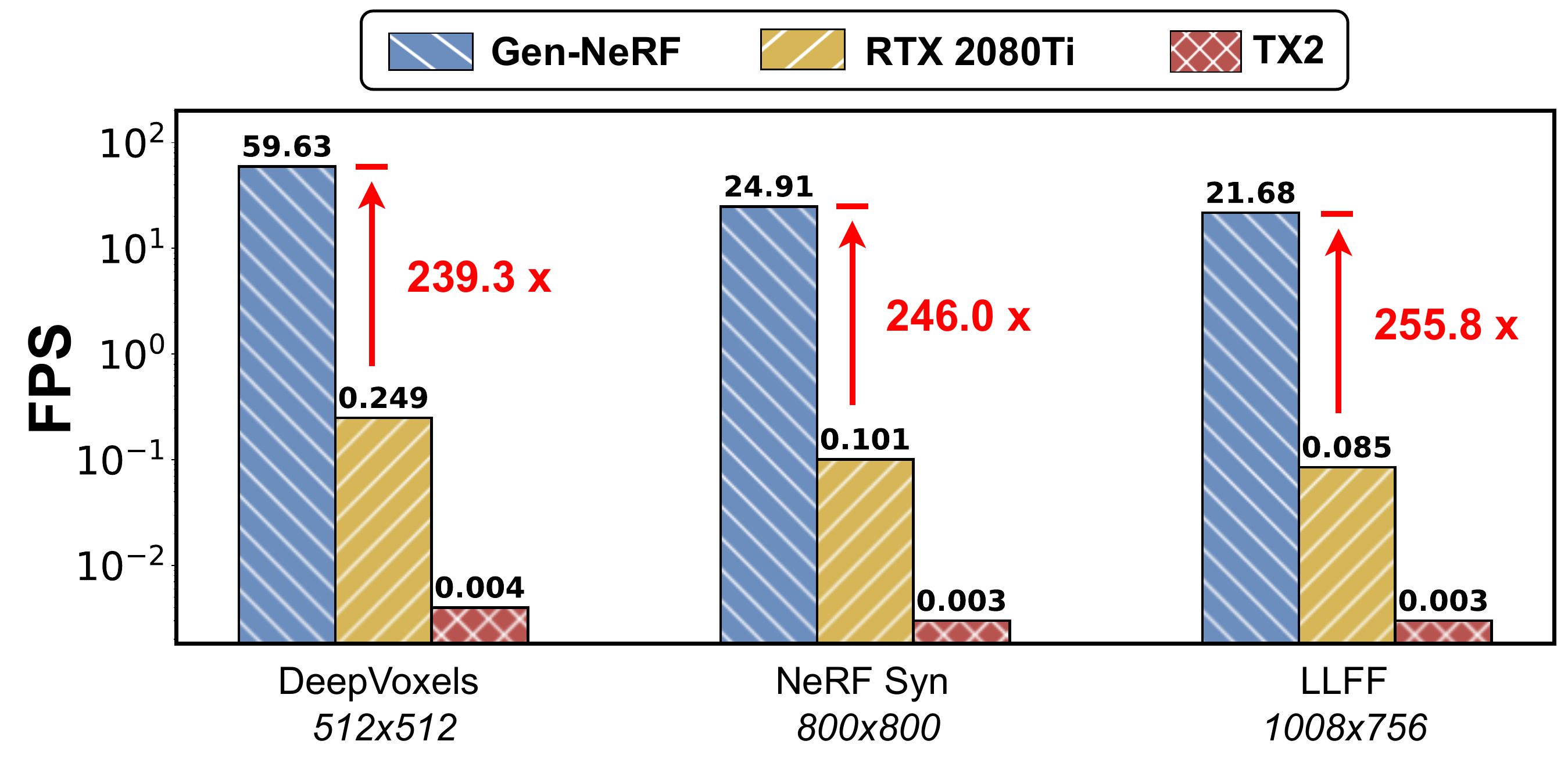}
\vspace{-1.2em}
\caption{Benchmark our \METHOD{} with two GPU devices on three datasets with different resolutions.}
\label{fig:throughput}
\vspace{-1em}
\end{figure}

\textbf{Benchmark with commercial GPUs.} We first benchmark our \METHOD{}'s accelerator with NVIDIA RTX 2080Ti desktop GPU and Jetson TX2 edge GPU for accelerating our \METHOD{}'s algorithm on three different datasets featuring different resolutions. The achieved throughput is shown in Fig.~\ref{fig:throughput}, where we can observe that (1) our \METHOD{}'s accelerator consistently outperforms both GPUs in throughput across all datasets, e.g., a 255.8$\times$/7448.9$\times$ FPS over RTX 2080Ti and TX2 on LLFF, respectively; (2) our accelerator can satisfy the real-time requirement ($\geqslant$ 24 FPS)~\cite{shimobaba2008real} for rending an 800$\times$800 image, indicating that our \METHOD{} is the first to enable real-time generalizable NeRFs with decent rendering quality, offering a promising NeRF solution for AR/VR applications.

\begin{figure}[!t]
\centering
\includegraphics[width=0.95\linewidth]{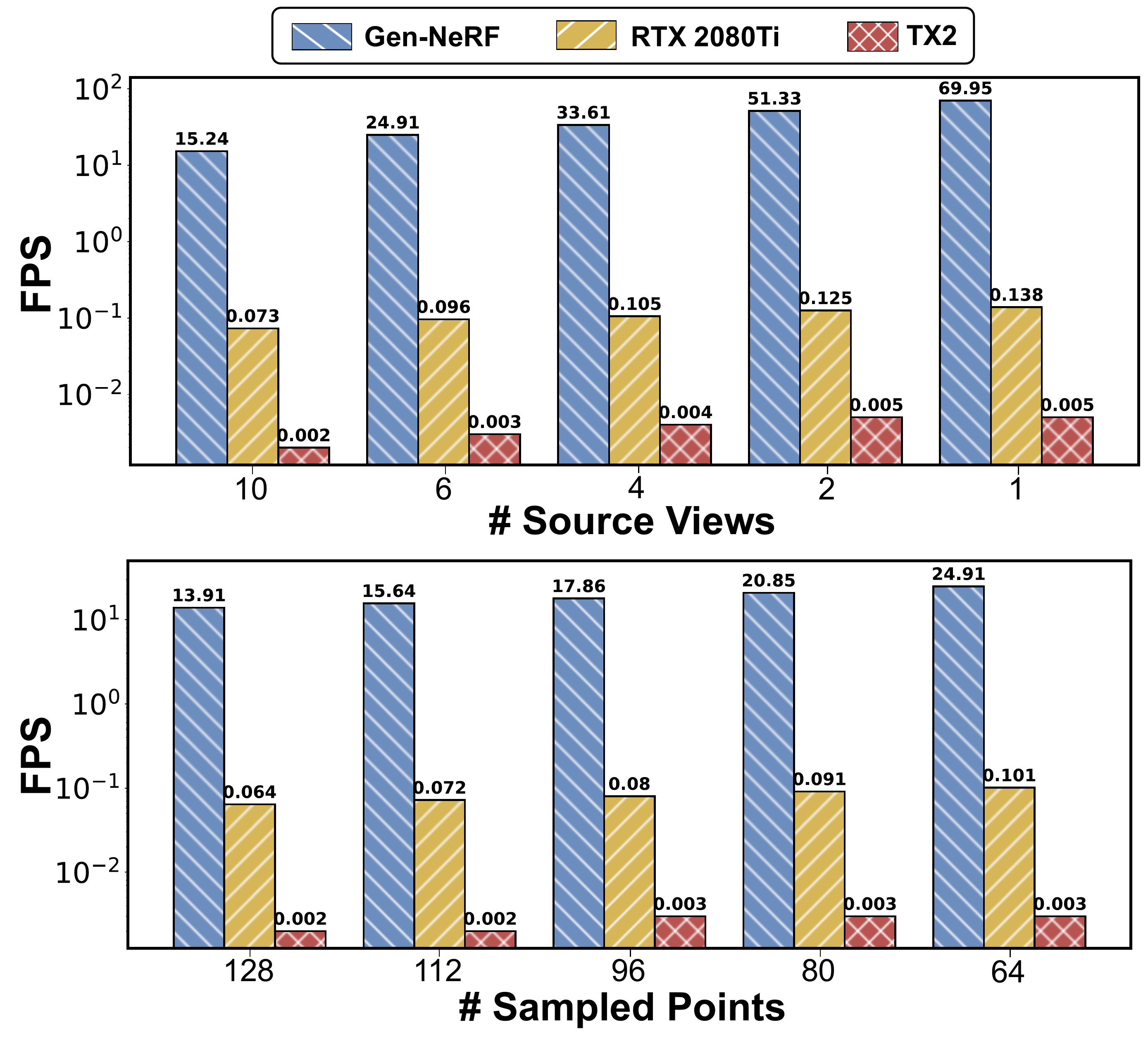}
\vspace{-1.2em}
\caption{Benchmark our \METHOD{} with two GPU devices on the NeRF Synthetic dataset with a resolution of 800$\times$800.}
\label{fig:scalability}
\vspace{-1.5em}
\end{figure}

\textbf{Evaluate the scalability with varied numbers of source views and sampled points in the focused sampling.}
As shown in Fig.~\ref{fig:scalability}, we can observe that our \METHOD{}'s accelerator consistently outperforms the two GPU baselines with $\geqslant$ 208.8$\times$ speed-up, indicating its scalability to different scenes with diverse source views and complexity for different use cases.

\textbf{Benchmark with SOTA NeRF accelerators.} 
We benchmark with ICARUS~\cite{rao2022icarus} in Tab.~\ref{tab:hardware_spec}, where the performance of ICARUS is their reported one. We can observe that \METHOD{} can outperform ICARUS with $>$1000$\times$ FPS under a comparable area, indicating that our \METHOD{}, featuring an algorithm-hardware co-design, is a more promising solution for NeRF acceleration.

\begin{table}[t!]
\caption{Specifications of our Gen-NeRF and baseline accelerators.}
\centering
\vspace{-1em}
\setlength{\tabcolsep}{2pt}
  \resizebox{1\linewidth}{!}
  {
    \begin{tabular}{c|c|ccc}
    \toprule
    \textbf{Device} & \textbf{Gen-NeRF} & \ \textbf{ICARUS~\cite{rao2022icarus}} & \ \textbf{Jetson TX2~\cite{tx2}} & \ \textbf{Nvidia RTX 2080Ti~\cite{2080ti}} \\
    
    \midrule
    SRAM  & 0.8 MB & 0.96 MB & 2.5MB & 29.5MB\\
    
    \midrule
    Area  & 17.80 mm$^2$ & 16.5 mm$^2$ & 350 mm$^2$ & 754 mm$^2$\\
    
    \midrule
    Frequency   & 1.0 GHz & 400 MHz &0.9 GHz & 1.35GHz\\
    
    \midrule
    DRAM &   \ LPDDR4-2400\,\,   & - & LPDDR4-1600 & GDDR6 \\
    Bandwidth &  17.8 GB/s  & - & 25.6 GB/s & 616GB/s \\
    
    \midrule
    Technology  & 28 nm & 40 nm & 16nm & 12nm \\ 

    \midrule
    Typical Power & 9.7 W & 282.8 mW & 10 W &250W \\
    
    \midrule
    Typical FPS & 24.9 & 0.02 & 0.003 & 0.096 \\
    
    \bottomrule
    \end{tabular}
    }
  \label{tab:hardware_spec}
  \vspace{-0.5em}
\end{table}

\textbf{Ablation study on the dataflow and feature storage format.}
To validate the effectiveness of our proposed dataflow and spatial interleaving feature storage format, we benchmark \METHOD{} with its three variants as baselines, including \textit{Var-1} w/o our dataflow, which instead prefetches and processes a patch sliced along the row dimension and column dimension ($W$ and $H$ in Fig.~\ref{fig:dataflow}) with a constant patch size $\{\delta h_{opt}, \delta w_{opt}, \delta d_{opt}\} = \{k,k,D\}$, where $k$ is the maximal value that satisfies the prefetch buffer size; \textit{Var-2} w/o both our dataflow and feature storage format, which instead stores scene features in DRAM and SRAM via a row-wise interleaving manner as shown in Fig.~\ref{fig:storage} (a) on top of \textit{Var-1};  \textit{Var-3}, which uses a view-wise interleaving manner on top of \textit{Var-1}.

\begin{figure}[t]
\centering
\includegraphics[width=\linewidth]{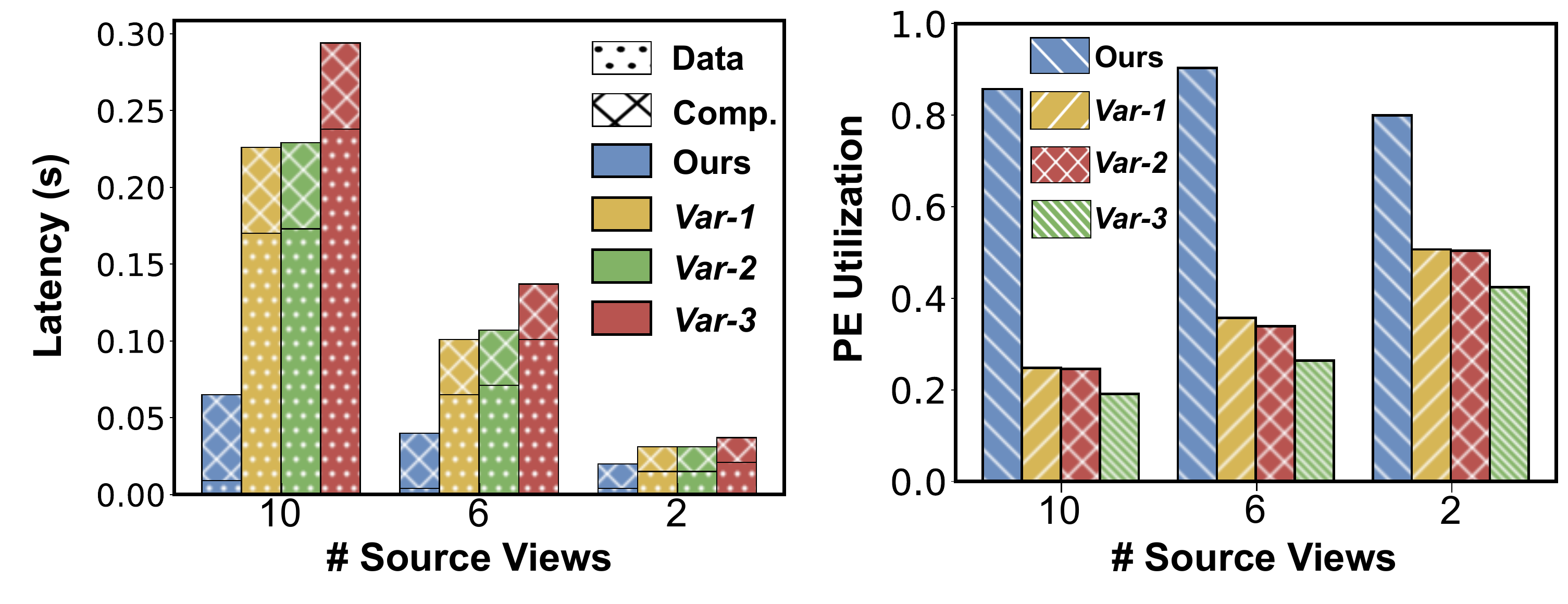}
\vspace{-2.5em}
\caption{Visualize the latency breakdown and utilization of \METHOD{}'s variants.}
\label{fig:ablation}
\vspace{-1.7em}
\end{figure}

We show the latency breakdown of data movement and compute and the resulting PE utilization under different numbers of source views in Fig.~\ref{fig:ablation}. We can observe that (1) the overall latency of \textit{Var-1} is bounded by memory access, since the latency of data movement is larger than that of compute in the pipeline,  resulting in low PE utilization. In contrast, our \METHOD{} successfully hides the data movement latency behind the compute time, indicating the effectiveness of our dataflow in enhancing scene feature reuse; and (2) as compared to \textit{Var-1}, \textit{Var-2} and \textit{Var-3} is even more memory-bounded due to the unbalanced communication volume to different memory banks, indicating the effectiveness of our spatial interleaving strategy in avoiding bank conflicts.

\section{Related Work}
\label{sec:related_work}

\textbf{View synthesis and NeRF.}
The task of view synthesis aims to render photorealistic images from novel viewpoints based on observed images of an object or scene.
To support free-viewpoint rendering from sparsely sampled images, explicitly or implicitly reconstructing the 3D representations of the objects/scenes is typically required. 
NeRF~\cite{mildenhall2020nerf} has gained increasing popularity thanks to its implicit scene representation, which fits each scene as a continuous 5D radiance field parameterized by an MLP and does not suffer from the aforementioned drawbacks. Follow-up works extend NeRF for generative modeling~\cite{chan2020pi, schwarz2020graf}, dynamic scenes~\cite{li2021, ost2020neural,park2021nerfies, park2021hypernerf, pumarola2021d}, or lighting and reflection modeling~\cite{bi2020, nerv2021, verbin2022ref}.

\textbf{Generalizable NeRFs.} To avoid per-scene optimization and enable the cross-scene generalization capability of NeRF, generalizable NeRF variants~\cite{yu2021pixelnerf,wang2021ibrnet,chen2021mvsnerf,reizenstein2021common,liu2022neural, johari2022geonerf, wang2022attention, xu2022point} are proposed to train cross-scene multi-view aggregators, which reconstruct the radiance field of a new scene via a one-shot forward pass. In particular,~\cite{yu2021pixelnerf,wang2021ibrnet,reizenstein2021common} condition NeRF on a set of source views from the new scene via feeding the extracted scene features from the source views into NeRF.~\cite{chen2021mvsnerf} extracts the scene features leveraging plane-swept cost volumes, which are widely used in multi-view stereo, for enhancing geometry awareness.~\cite{liu2022neural} further predicts the visibility of 3D points to each source view to avoid inconsistent features from invisible views. 
However, generalizable NeRFs pose new challenges for NeRF acceleration as analyzed in Sec.~\ref{sec:hardware}, calling for new efficiency-enhancing techniques.

\textbf{NeRF accelerators.}
As NeRF is still an emergent field, limited attempts have been made for NeRF acceleration from the accelerator perspective. ICARUS~\cite{rao2022icarus} proposes an architecture for the vanilla MLP-dominated NeRF~\cite{mildenhall2020nerf}, which accelerates each component of NeRF via customized modules;~\cite{zheng2022rram} develops a resistive random access memory (RRAM)-based NeRF accelerator based on the parallel nature of NeRFs' rendering process to enhance the utilization of the RRAM array. 
However, both of the aforementioned works (1) assume different rays can be executed in parallel without acquiring extra features stored in DRAM and thus their proposed accelerators cannot well handle the data movement cost in generalizable NeRFs, and (2) lack algorithmic improvements to exploit the intrinsic redundancy in NeRF.
\cite{li2022rt} is a pioneering work that features an algorithm-hardware co-design for NeRFs, which tackles the latency bottlenecks, caused by querying the occupancy grid and calculating the voxel-wise feature embeddings, of an efficient NeRF design dubbed TensoRF~\cite{chen2022tensorf}. 
Nevertheless, their techniques only focus on a specific type of NeRFs, i.e., TensoRF~\cite{chen2022tensorf}, which lacks cross-scene generalization capability because of its use of scene-specific occupancy grid and voxel-wise embeddings. Therefore, their acceleration methods are not applicable to generalizable NeRFs.
In contrast, our work is the first algorithm-hardware co-design targeting generalizable NeRFs, featuring a win-win in both rendering efficiency and cross-scene generalization, both of which are crucial for real-device deployments of NeRFs in AR/VR applications. 
In addition, instead of targeting one specific kind of NeRFs, our delivered techniques and insights can be generally applicable to generalizable NeRF variants~\cite{yu2021pixelnerf,wang2021ibrnet,chen2021mvsnerf,liu2022neural,reizenstein2021common}.

\vspace{-0.5em}
\section{Conclusion}
\label{sec:conclusion}

To enable efficient and generalizable NeRFs towards real-time novel view synthesis in AR/VR applications, we propose \METHOD{}, which is the first algorithm-hardware co-design framework dedicated to accelerating generalizable NeRFs. 
Our \METHOD{} identifies the unique opportunities for generalizable NeRF acceleration, including the diverse sparsity distributions in a 3D space and the scene feature reuse opportunities derived from the epipolar geometric relationship among different points and rays, leveraging which we develop a dedicated algorithm and accelerator, respectively, to push forward the achievable accuracy-efficiency trade-off of generalizable NeRFs.


\vspace{-0.5em}
\section*{Acknowledgement}
The work is supported by the National Science Foundation (NSF) through two CCF programs (Award numbers: 2211815 and 2312758) and CoCoSys, one of the seven centers in JUMP 2.0, a Semiconductor Research Corporation (SRC) program sponsored by DARPA.

\bibliographystyle{ACM-Reference-Format}
\balance
\bibliography{ref}

\end{document}